\PassOptionsToPackage{table}{xcolor}
\documentclass[]{xiaomi}

\usepackage[utf8]{inputenc}
\usepackage[T1]{fontenc}
\usepackage{hyperref}
\usepackage{url}
\usepackage{amsfonts}
\usepackage{nicefrac}
\usepackage{microtype}
\usepackage{xcolor}
\usepackage{overpic}
\usepackage{graphicx}
\usepackage{float}
\usepackage{array}
\usepackage{tabularx} 
\usepackage{makecell} 
\usepackage{pifont}
\usepackage{textcomp}
\usepackage{booktabs}
\usepackage{multirow}
\usepackage{subcaption}
\usepackage{amsmath}
\usepackage{amssymb}
\usepackage{ragged2e}
\usepackage{enumitem}
\usepackage{wrapfig}
\usepackage{xspace}
\usepackage{titlesec}
\usepackage{pgfplots}
\pgfplotsset{compat=1.18}

\usepackage{booktabs}
\usepackage{tabularx}
\usepackage{subcaption} 
\usepackage{threeparttable}

\titleformat{\paragraph}[runin]{\normalfont\bfseries}{\theparagraph}{0.5em}{}[\;]
\titlespacing*{\paragraph}{0pt}{0.8\baselineskip}{0.8em}

\newcommand{\figref}[1]{Fig.~\ref{#1}}
\newcommand{\tabref}[1]{Tab.~\ref{#1}}

\newcommand{\secref}[1]{Sec.~\ref{#1}}

\def\eg{\emph{e.g.}}

\title{Xiaomi Auto World Model 2.0: Causal World Models for Real-World Driving and Embodied Intelligence}

\affiliation{\hyperref[sec:contributors]{Xiaomi Auto World Model Team}}

\abstract{%

We present \textbf{Xiaomi Auto World Model 2.0}, a unified streaming causal world model for Physical AI. 
The model is built around two core capabilities: modeling and generating causal physical evolution in the real world, and unifying heterogeneous conditioning inputs within a shared framework for view-consistent generation.
Concretely, our architecture integrates View-Temporal Attention (VTA) to capture cross-view spatial dependencies and temporal dynamics, together with Multi-view Rotary Position Encoding (RoPE) to encode positional relationships across views and time.
We further introduce WorldRec 2.0, which reconstructs static and dynamic 3D Gaussians from multi-view, multi-temporal observations in a single forward pass and uses their renderings as explicit scene context to preserve geometry and appearance during long-horizon generation.
Heterogeneous conditioning inputs are fused through VAE-based latent feature concatenation within a shared representation space.
We incorporate Gated Linear Attention (GLA) into the temporal layers of VTA for efficient streaming inference, reducing computational cost and KV cache memory usage during multi-view rollouts.
To strengthen physical-dynamics modeling, we develop a physical-interaction data pipeline covering diverse interactions and long-tailed events.
From Pre-train onward, these data are combined with large-scale videos and heterogeneous visual modalities to learn visual, geometric, and dynamic priors.
During Mid-train, quality-aware mixed training across diverse view configurations further improves the model's ability to capture how actions and object states shape subsequent scene evolution.
Following distillation-based Post-train, we apply the resulting model to closed-loop simulation, closed-loop training, and World Action Model (WAM) applications.
These applications demonstrate its ability to generate physically consistent future states and support world-state prediction, policy learning, and action generation for Physical AI.
}

\date{November 2026}
\project{https://xiaomiautoworldmodel.github.io/CausalWorldModel/}

\begin{document}
\thispagestyle{firstheader}
\maketitle

\vspace{-3mm}
\begin{figure}[H]
  \centering
  \setlength{\abovecaptionskip}{3pt}
  \setlength{\belowcaptionskip}{0pt}
  \includegraphics[width=\linewidth]{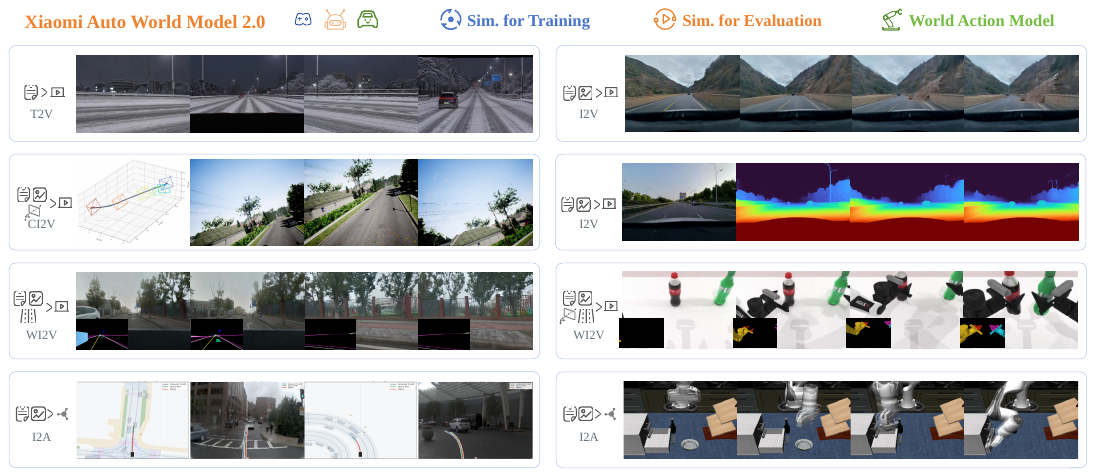}
  \caption{
  \textbf{Xiaomi Auto World Model 2.0} supports unified, streaming causal world modeling across general video generation, autonomous driving, and embodied robotics. 
  Examples include text-to-video and image-to-video generation, controllable prediction across varying views, and action-conditioned rollouts for embodied manipulation.
  The resulting world model can serve as a simulator for both training and evaluation, and as a component in a WAM.
  }
  \label{fig:teaser}
\end{figure}

\section{Introduction}
\label{sec:intro}

World models~\cite{XiaomiAutoWorldModel1, lingbotworld, hunyunavideo, dreamxworld, wan, Cosmos3, solarwm,wu2026video,zhu2026sanawm,xu2026wonder,yang2026neoverse,hu2026geometry,l2r,hu2026omni,l2r} aim to generate the evolution of a scene at future time steps based on environmental observations, actions, or task instructions. 
For Physical AI agents such as autonomous vehicles and embodied robots, this capability is fundamental to anticipatory decision-making and closed-loop control. 
Before changing lanes, avoiding obstacles, or performing emergency braking, a vehicle must anticipate the motion of surrounding traffic participants and its potential consequences. 
Similarly, before grasping, pushing, or placing an object, a robot must infer how its actions will change the state of the object and the spatial configuration of the scene. 
Therefore, an effective world model should do more than generate visually plausible future frames. 
It should also capture the relationships among actions, object states, and environmental changes, and predict the physical consequences induced by different control conditions.


Recent video world models have made substantial progress in predicting and generating future dynamics, with growing applications in autonomous driving~\cite{drivedreamer, drivewm, vista, cosmosdrive, horizondrive, OmniDreams, XWorld, magicdriveV2, guan2026cdcp, lu2026xiaomi,luo2026reflect,zhu2026worldsplat,ji2025cogen,jin2025posepilot,genesis,dream4drive,phygenesis} and embodied intelligence~\cite{wu2023daydreamer, causalwm, gao2026dreamdojo, bi2026motus, ye2026world, xu2026world, xu2026univiewvla}. 
Nevertheless, as summarized in \tabref{tab:method-comparison}, existing models provide limited unified support for various domains, view configurations, control inputs, and downstream applications. 
Building a general-purpose world model for Physical AI therefore requires addressing two fundamental challenges. First, physical interactions in the real-world follow a highly long-tailed distribution~\cite{wang2025alpamayo, collaboration2023open}. 
Naturally collected videos are dominated by routine and frequently occurring events, providing limited coverage of rare but safety-critical interactions such as collisions, emergency evasive maneuvers, and object failures. 
Scaling conventional video data alone is therefore insufficient for learning the diverse physical consequences. 
Second, Physical AI systems operate under heterogeneous observation and conditioning configurations, ranging from single-camera inputs to multi-camera systems with different layouts and scene contexts~\cite{gaia2, team2024octo,tan2026ufo}. 
Models designed for fixed view configurations and conditioning formats generalize poorly across scenarios.
Meanwhile, scaling to more views and longer temporal contexts substantially increases computational and memory requirements, limiting long-horizon closed-loop generation. 
Our goals are:


\definecolor{lightpurple}{HTML}{F2F0FF}
\definecolor{purpleframe}{HTML}{D8D3F2}

\begin{tcolorbox}[
  enhanced,
  breakable,
  colback=lightpurple,
  colframe=purpleframe,
  boxrule=0.6pt,
  arc=0mm,
  left=8pt,
  right=8pt,
  top=6pt,
  bottom=6pt
]
\fontfamily{phv}\selectfont

\textbf{Goal 1. Model and generate the causal evolution in the real world.}

\medskip

\textbf{Goal 2. Unify heterogeneous conditioning across variable-view configurations.}

\medskip

\textbf{Goal 3. Enable diverse downstream Physical AI applications.}

\end{tcolorbox}


We therefore present \textbf{Xiaomi Auto World Model 2.0}, a unified streaming causal world model for Physical AI. 
Fig.~\ref{fig:teaser} highlights the model's unified capabilities across general video generation, autonomous driving, and embodied robotics.
With variable-view configurations and heterogeneous conditioning inputs, our model enables controllable generation for simulation-based training and evaluation, as well as WAM applications~\cite{li2026drivevla,xia2026reworld}.


To improve physical-dynamics modeling, we construct a physical-interaction data pipeline that integrates real-world driving and robotics data with physically causal data.
The physically causal subset draws on simulator engines, video generators, and public media, covering physically grounded interactions and long-tailed events.
From the Pre-Train stage onward, these data are jointly trained with large-scale general-purpose videos and diverse visual representations, enabling the model to acquire general visual and dynamic priors, strengthen its geometric understanding, and form an initial representation of physical evolution.
During Mid-Train, quality-aware training on high-quality data spanning diverse view configurations further strengthens the modeling of how actions and object states shape subsequent scene evolution, improving the representation and generation of real-world physical interactions and long-tailed dynamics while preserving generalization across scenes and observation settings.


We introduce a unified generation framework for heterogeneous conditioning inputs and variable-view future-scene generation in a shared latent space.
View-Temporal Attention (VTA) captures cross-view dependencies and temporal dynamics across varying view configurations, while Multi-view Rotary Position Encoding (RoPE) encodes positional relationships across views and time.
Heterogeneous conditioning inputs are mapped into compatible latent representations and fused through VAE-based feature concatenation.
Through this interface, we incorporate WorldRec 2.0 as explicit 3D scene memory grounded in real observations.
It reconstructs static and dynamic 3D Gaussians from multi-camera, multi-temporal observations in a single forward pass, using scene-structured registers for efficient information exchange among point-aligned queries.
Renderings at target camera poses and timestamps provide geometry-aware priors that anchor geometry, appearance, and cross-view correspondence in observed regions, while the generator completes appearance and motion details beyond the observations.
Together, these components support single-view, autonomous-driving multi-view, and embodied multi-view generation within a shared architecture, with flexible conditioning and consistent scene representations across views and time.


Building on these designs, we develop a unified causal world model that supports variable-view generation and flexible conditioning.
To improve the efficiency of long-horizon multi-view rollouts, we incorporate Gated Linear Attention (GLA) into the temporal attention layers of VTA, reducing the computational complexity and the memory overhead associated with state caching during inference. 
Distillation-based post-training further improves streaming efficiency and supports stable generation over extended sequences. 
This reduces the computational and memory costs of multi-view rollouts, allowing the model to continuously generate future scenes and extend short-term prediction to streaming world simulation over longer time horizons. 
We evaluate the Post-trained model across three downstream settings. 
First, for simulation-based evaluation, we assess its general generation and world-modeling capabilities through general video generation on PAI-Bench-G~\cite{paibench}, synchronized multi-view generation under long-tailed and physically causal scenarios, scene perception via depth prediction and referring video segmentation, and camera-pose control on WorldScore~\cite{worldscore}, TartanAir~\cite{wang2020tartanair}, and RealCam-Vid~\cite{zheng2025realcam}. 
We further use the model as a controllable world simulator to evaluate multi-view driving generation on nuScenes~\cite{Caesar2019nuScenesAM}, embodied interaction generation on TriWorldBench~\cite{triworldbench2026}, and long-horizon driving simulation on internal data. 
Second, for simulation-based training, the model generates future states and interaction feedback in closed-loop rollouts to support policy learning. 
We evaluate the resulting policy on NAVSIM~\cite{dauner2024navsim}, using 1000 representative test scenes.
Finally, for WAM applications, we transfer the learned visual-dynamics and physical-interaction representations to downstream decision-making and control, evaluating autonomous-driving planning on NAVSIM~\cite{dauner2024navsim}, embodied manipulation on LIBERO Plus~\cite{Liberoplus}.
We additionally evaluate WorldRec 2.0 on the Waymo Open Dataset~\cite{xu2026wod}, use nuScenes~\cite{Caesar2019nuScenesAM} and Argoverse 2~\cite{Wilson2023Argoverse2} for cross-dataset evaluation, while reporting qualitative results on the internal MiData dataset.
Together, these experiments evaluate the model's generation quality, physical and geometric consistency, controllability, simulation capability, and downstream utility across diverse Physical AI scenarios.


\begin{table}[t]
  \centering
  \caption{Comparison of recent world models in terms of generation horizon, camera number, control, domain, downstream usage, and real-time. \emph{Cam.} denotes camera pose; \emph{action} denotes action sequences (including robot action sequences and WASD inputs); \emph{Interact.} denotes open-interaction world models; \emph{Sim.} denotes closed-loop simulation or training; \emph{policy} indicates whether future action sequences are predicted.}
  \label{tab:method-comparison}
  \small
  \setlength{\tabcolsep}{4pt}
  \renewcommand{\arraystretch}{1.25}
  \begin{tabular*}{\linewidth}{@{}l|@{\extracolsep{\fill}}cccccc@{}}
    \toprule
    Method &
    \makecell[c]{Generation\\Duration} &
    \makecell[c]{Camera\\views} &
    Control &
    Domain &
    Applications &
    Real-time \\
    \midrule
    \makecell[l]{LingBotWorld \cite{lingbotworld}} & Minutes & 1V & \makecell[c]{Cam.,action} & General & Interact. & \checkmark \\
    \makecell[l]{DreamXWorld \cite{dreamxworld}} & Minutes & 1V & \makecell[c]{Cam.,action} & General & Interact. & \checkmark \\
    CausalWM\cite{causalwm} & Seconds & 1--4V & Action & Robot & Sim. & -- \\
    \makecell[l]{HorizonDrive \cite{horizondrive}} & Minutes & 1V & Layout & Driving & Sim. & -- \\
    OmniDreams\cite{OmniDreams} & Minutes & 1/4/7V & Layout & Driving & Sim.,policy & \checkmark \\
    X-World\cite{XWorld} & $\sim$24\,s & 7V & Layout & Driving & Sim. & \checkmark \\
    Ours(1.0)\cite{XiaomiAutoWorldModel1} & Minutes & 1V/3V & Layout & Driving & Sim. & \checkmark \\
    \textbf{Ours(2.0)} & Minutes & 1--7V & \makecell[c]{Cam., action,\\layout} & Phys.\ AI & \makecell[c]{Interact.,Sim.,\\policy} & \checkmark \\
    \bottomrule
  \end{tabular*}
\end{table}
\section{Data}
\label{sec:data}

\begin{figure}[h]
    \centering
    \includegraphics[width=1.0\linewidth]{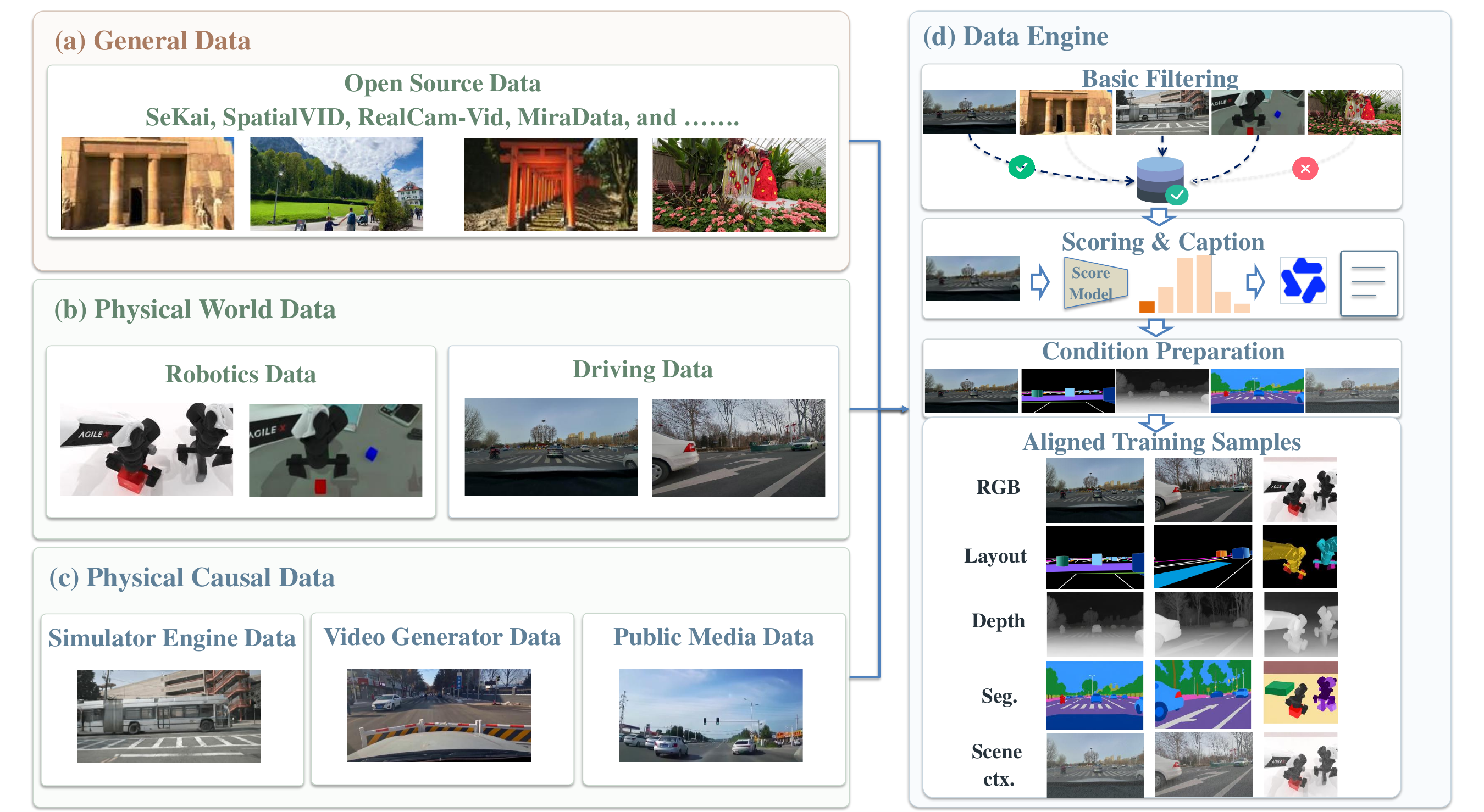}
    \caption{\textbf{Overview of our data pipeline.} We integrate three data categories: general data, physical world driving and robotics data, and physical causal data. The collected videos undergo basic filtering, scoring, and captioning. We further prepare world-control conditions for the corresponding data subsets, including layouts, and scene contexts, to support causal world model training.}
    \label{fig:data_pipeline}
\end{figure}

Data are essential for model training. To train a causal world model that respects physical causal laws, we built multi-view datasets that emphasize physical causality.
We partition data sources into three categories: \textit{(i) general data}, comprising camera, text, and video data; \textit{(ii) physical-world data}, comprising autonomous-driving and robotics data; and \textit{(iii) physical-casual data}.
General data provide broad priors on visual appearance, geometric structure, language, and motion patterns. Physical-world data ground these priors in embodied observations and actions. Physical-causal data further include safety-critical corner cases and counterfactual trajectories that are underrepresented in naturally collected data. 
All data modalities are subjected to a unified cleaning and re-annotation pipeline. For high-quality physical world data, we also estimate associated multi-modal conditional annotations, including geometric representations, segmentation masks, and scene layout attributes.
The following sections introduce data sources and their curation pipelines, followed by data filtering, re-captioning, and condition annotation creation, as summarized in \figref{fig:data_pipeline}.

\subsection{General Data} 
\label{sec:data-pretrain}

Before domain-specific specialization, we use large-scale general data to learn priors about visual appearance, language, motion dynamics, and scene structure. These data are organized into camera, text, and video sources and are independent of the downstream driving and robotics domains.

Leveraging the broad availability of video data, we curate raw videos and superimpose high-fidelity multimodal annotations. To jointly optimize dataset coverage and experimental controllability, we integrate (i) internet-scale web videos to capture long-tail variation in scenes, objects, and motion patterns, and (ii) structured open-source datasets that provide reliable metadata and support evaluation under more stringent, domain-specific conditions.

\paragraph{Web-crawled Data.} We follow established collection protocols used in recent video foundation models. Specifically, we adopt Pandas 70M~\cite{chen2024panda} as the primary source and apply strict filtering criteria to ensure minimum visual quality and temporal consistency. In addition to their scale, these videos often include textual descriptions that provide weak semantic supervision, thereby exposing the model to a broad distribution of everyday events and dynamics.

\paragraph{Open-source Data.} We target datasets that provide explicit supervisory signals that are typically absent in unconstrained web content, including camera poses, geometry-related conditions, and driving or embodiment metadata. These data sources focus on operationally grounded domains, such as egocentric, autonomous driving, and robotic manipulation, and thus complement the breadth of web video with well-defined constraints. Concretely, we aggregate video data from SpatialVID-HQ~\cite{wang2026spatialvid}, Sekai~\cite{li2026sekai}, RealCam-Vid~\cite{zheng2025realcam}, and TartanAir~\cite{wang2020tartanair}. 


\subsection{Physical World Data}
\label{sec:data-physical}

Physical-world data connect visual observations to embodied states, actions, and consequences. They comprise autonomous-driving data and robotics data. 







\subsubsection{Driving Data}
The autonomous-driving data provide multi-view camera streams and vehicle-centric signals, including driving trajectories, maps, and sensor metadata. Specifically, we use PhysicalAI-Autonomous-Vehicles~\cite{wang2025alpamayo}, nuPlan~\cite{caesar2021nuplan}, nuScene~\cite{Caesar2019nuScenesAM}, and our self-collected MiData. PhysicalAI-Autonomous-Vehicles captures a broad range of traffic densities, weather conditions, obstacles, and interactions with vulnerable road users. Its sensor suite includes synchronized multi-camera and LiDAR recordings for all clips, with radar data available for approximately half of the clips. MiData is primarily collected by our data-acquisition fleet and spans heterogeneous operational design domains. Spatially, it includes urban and rural roads, highways, underground garages, and parking lots. Temporally, it covers different times of day and multiple seasons. Each MiData sequence contains 11 synchronized camera streams recorded at 30 Hz and one LiDAR stream, together with ego-vehicle state measurements. Sequences are partitioned into approximately 30 s clips, and a cloud-based automatic labeling system is applied to produce core scene annotations, including object detection, lane-boundary detection, traffic-sign recognition and traffic-light recognition.

\subsubsection{Robotics Data}
Robotics data cover egocentric observations and robot-manipulation episodes, together with actions, proprioception, object states, and language instructions when available. We include datasets used for vision-language-action training, such as AgiBot World Beta~\cite{bu2025agibot}, DROID~\cite{Khazatsky2024DROIDAL}, EgoDex~\cite{hoque2026egodex}, Galaxea Open-World Dataset~\cite{jiang2026galaxy}, InternData-A1~\cite{tian2026interndata}, Open X-Embodiment~\cite{collaboration2023open}, RoboCOIN~\cite{Wu2025RoboCOINAO}, RoboMIND~\cite{Wu2024RoboMINDBO}, and RoboTwin 2.0~\cite{Chen2025RoboTwin2A}. Collectively, these datasets span real-world and simulated collections, single-arm and bimanual embodiments, tabletop and open-world scenes, and both human egocentric and robot-centric views. They provide large-scale heterogeneous supervision, ranging from tens of thousands to millions of trajectories, thousands of hours of video, and hundreds of tasks, skills, and scenes, with language annotations where available. We align their observations and controls with the same temporal and geometric annotation interface. These data extend the physical-world category from road interactions to contact-rich manipulation and support transfer to embodied tasks.

\subsection{Physical Causal Data}

Although general data and physical world data exhibit substantial diversity, they frequently underrepresent emergency events that occur in the physical world, including vehicular collisions and other rare or highly specialized incidents. To mitigate this limitation, we curate (i) simulator engine data and (ii) video generator data to construct these corner-cases and counterfactual simulation.

\label{sec:data-physical-causal}

\paragraph{Simulator Engine Data.}

To address the limited prevalence of extreme physical interactions in real-world driving corpora, such as collisions, roadway departures, and aggressive maneuvers, we construct a large-scale, multi-view synthetic dataset using the simulator engine, including: Blender~\cite{blender2025}, UE5~\cite{unrealengine5}, and CARLA~\cite{dosovitskiy2017carla}, following \cite{phygenesis}. Blender and UE5 is used to generate higher-quality multi-view video and geometric conditions. CARLA is a high-fidelity autonomous-driving simulation platform that supports controlled generation of road networks, vehicles, traffic participants, and diverse environmental conditions, including weather. We build upon the route configurations provided by Bench2Drive~\cite{jia2024bench2drive} and introduce localized behavioral perturbations while keeping the underlying maps, traffic participants, weather settings, and background traffic behavior unchanged. This protocol yields two subsets: CARLA Ego and CARLA Adv. In CARLA Ego, we directly perturb the behavior of the ego vehicle. In CARLA Adv, we instantiate an additional vehicle near the ego vehicle and perturb its behavior, such that the resulting physical challenges primarily arise from interactions between the two vehicles.
Although the data produced by physics simulators exhibit good physical consistency, they are limited to specific traffic scenarios, thereby restricting diversity. To address this, we employ advanced video generators for data simulation.

\paragraph{Video Generator Data.}

To complement the scenario coverage of physics-based simulation, we construct a long-tail driving dataset with open-sourced video diffusion model. Starting from 10,000 unique forward-facing driving clips and their scene descriptions, we independently sample road environments, traffic conditions, weather and illumination, visibility constraints, hazardous objects, event progression, and ego-vehicle responses, and combine them into structured generation prompts. We use three complementary generation modes: T2VA provides text-only scene conditioning, I2VA uses the first frame of the source clip to preserve the initial scene layout and camera view, and Ref2VA uses a 1s reference sequence to retain scene structure, traffic motion, and camera dynamics while introducing new hazards or physical interactions. All prompts are reviewed for semantic and physical consistency, including weather–illumination compatibility, road structure, object motion, occlusion causality, vehicle reaction time, braking distance, and temporal continuity. The prompts that fail these checks are repaired before generation. 
Before using them for training, we manually review the generated videos and filter out low-quality or inconsistent samples.

\subsection{Data Engine} 

After collecting the aforementioned data, we proceed quality filtering, scoring, captioning, and condition preparation.


\paragraph{Initial Quality Filtering.}
We first apply a coarse filtering stage to exclude videos that are technically invalid or exhibit obvious visual degradation. Web-crawled videos frequently have inconsistent or mislabeled metadata, poor visual quality, or undesirable overlays. Consequently, we screen samples using basic structural and visual criteria, including extreme aspect ratios, insufficient frame counts, and discrepancies between the observed visual content and the associated metadata. We also remove pseudo-high-definition videos, such as low-resolution videos upsampled to a higher nominal resolution or incorrectly annotated as high-resolution. Videos with excessive on-screen text, subtitles, watermarks, or other salient overlays are also discarded. Clips exhibiting severe blur, including Gaussian-like blurring artifacts, are further excluded. Moreover, each video is converted to the YUV color space, and clips with extreme luminance distributions are filtered out to eliminate markedly underexposed or overexposed content.


\paragraph{High-Quality Scoring.}
After the initial filtering stage, we apply a second-stage curation to retain video clips with high visual quality, sufficient motion, and stable temporal dynamics. We first assess perceptual quality from both aesthetic and imaging perspectives. Overall video quality is quantified using DOVER~\cite{wu2023exploring}, and clips with a DOVER score below $0.4$ are excluded. In addition, we evaluate imaging quality with COVER~\cite{he2024cover}, which targets degradations such as compression artifacts and transmission-induced distortions. Samples exhibiting pronounced image-quality impairments are removed.
Next, we characterize clip-level motion properties. We use the VMAF motion score~\cite{li2016toward} as a surrogate for global motion magnitude and retain only samples with scores in the range $[2,14]$. This criterion filters out near-static content with insufficient motion as well as clips dominated by excessively rapid or unstable motion. We further apply heuristic constraints on camera trajectory. Specifically, we discard clips that (i) contain multiple view reversals exceeding $150^\circ$ within a 10-second clip, (ii) exhibit inter-frame view-angle changes greater than $60^\circ$, or (iii) include abrupt camera translations whose short-term displacement exceeds five times the mean translation magnitude computed over the entire video.
Finally, we manually inspect the remaining clips to remove residual failure cases that are not reliably captured by automated metrics. This review emphasizes overall visual integrity and temporal coherence, thereby ensuring that the retained high-quality data are suitable for mid-train and post-train.




\paragraph{Captioning.} 
For the videos retained after quality curation, we apply a unified segmentation and re-captioning pipeline. We uniformly sample all videos at 16~Hz and divide them into non-overlapping clips of 93 frames, corresponding to approximately 5.8~s per clip. Clips at the end of a video containing fewer than 50 frames are discarded to avoid generating excessively short training samples.
Each resulting clip is then passed to Qwen3.8~\cite{qwen3.8} for automated re-captioning. We adopt a two-stage pipeline that first produces a structured scene analysis and then converts this analysis into a dense, temporally coherent narrative description. Given a sequence of video frames, Qwen3.8 first completes the \textit{Scene Analysis} stage and subsequently generates the \textit{Final Narrative} conditioned on the structured analysis. The two outputs are enclosed by the \texttt{<scene\_draft>} and \texttt{<final\_prompt>} tags, respectively.

In the first stage, Qwen3.8 examines the clip frame by frame and populates a predefined JSON schema describing its observable visual and temporal content. The structured representation captures the major subjects and their appearance, pose, location, orientation, actions, and state changes, together with the surrounding environment, lighting, aesthetics, cinematography, visual style, contextual information, textual elements, temporal segments, transitions, and audio content. Temporal annotations use a unified \texttt{M:SS-M} format, with timestamps measured relative to the beginning of the clip.
The annotation follows a strict observation-based policy: only visually or audibly observable information is recorded, and no unobserved details are inferred or hallucinated. Fields that cannot be determined are left empty. Basic video metadata, including resolution, aspect ratio, duration, and frame rate, is not generated by the model and is instead injected automatically from the source video. The resulting structured representation serves as a normalized intermediate description for the subsequent narrative generation stage.

In the second stage, Qwen3.8 rewrites the structured scene analysis into a dense narrative caption. The output is written as a single paragraph consisting of multiple coherent sentences, without lists or bullet points. The narrative is required to cover the major subjects, scene setting, temporally ordered actions, relevant background elements, lighting conditions, cinematographic characteristics, and temporal structure of the clip.
For each video clip, the re-captioning pipeline produces two complementary outputs: a structured JSON scene analysis and a dense narrative caption. The former provides machine-readable descriptions of visual attributes, temporal events, cinematography, and other structured information, while the latter provides a fluent and temporally coherent natural-language description. Together, these representations provide both structured supervision and dense textual descriptions.







\paragraph{Condition Preparation.}

Each training sample is paired with two types of conditions: a camera-view pose $p_i$ and a world control $w_i$. The world control has three forms: \emph{Driving layouts}, \emph{Robotic layouts}, and \emph{Scene contexts}. For the camera-view pose, we use the poses provided by structured datasets whenever available. For samples without camera metadata, we estimate the pose with DA3~\cite{lin2026depth}. We also include data collected in game engines and Unreal Engine, which provide high-quality camera poses and complement the estimated poses. \emph{Driving layouts} are built from 3D object detections, lane detections, traffic-light recognition, and traffic-sign recognition. We obtain these annotations through auto-labeling for our private MiData collection. Public datasets generally provide the corresponding labels, which we convert to a common format. We then render the aligned annotations with the Cosmos-Transfer tools~\cite{cosmosdrive}. \emph{Robotic layouts} are constructed primarily from the Robotwin dataset for training and evaluation on TriWorldBench~\cite{Liu2026TriWorldBenchAT}. Given the robot arm's action sequence and configuration, we render its state from the corresponding URDF model. \emph{Scene contexts} are designed to improve consistency in driving closed-loop simulation. For an input clip, the feed-forward reconstruction method in \secref{sec:worldrec2} first produces a dynamic 3D Gaussian Splatting (3DGS) scene. During simulation, each new trajectory is rendered from the corresponding novel views of this scene, and the resulting per-frame renders serve as world control so that the causal DiT can preserve the geometry and appearance of the original scene. During training, we create input-target pairs by degrading the ground-truth render at the original view. These degradations approximate the distribution of novel-view renders from dynamic 3DGS and are applied to a small subset of the driving data (5\%). We use three simple degradation types: object-support degradation reduces and irregularizes the opacity of tracked objects to mimic incomplete reconstruction, occlusion-cone holes remove parts of the static background that were hidden in the source views, and depth degradation perturbs depths and boundaries to mimic novel-view errors.

\begin{figure}[!t]
    \centering
        \begin{subfigure}[b]{0.48\textwidth}
        \centering
        \includegraphics[width=1.0\linewidth]{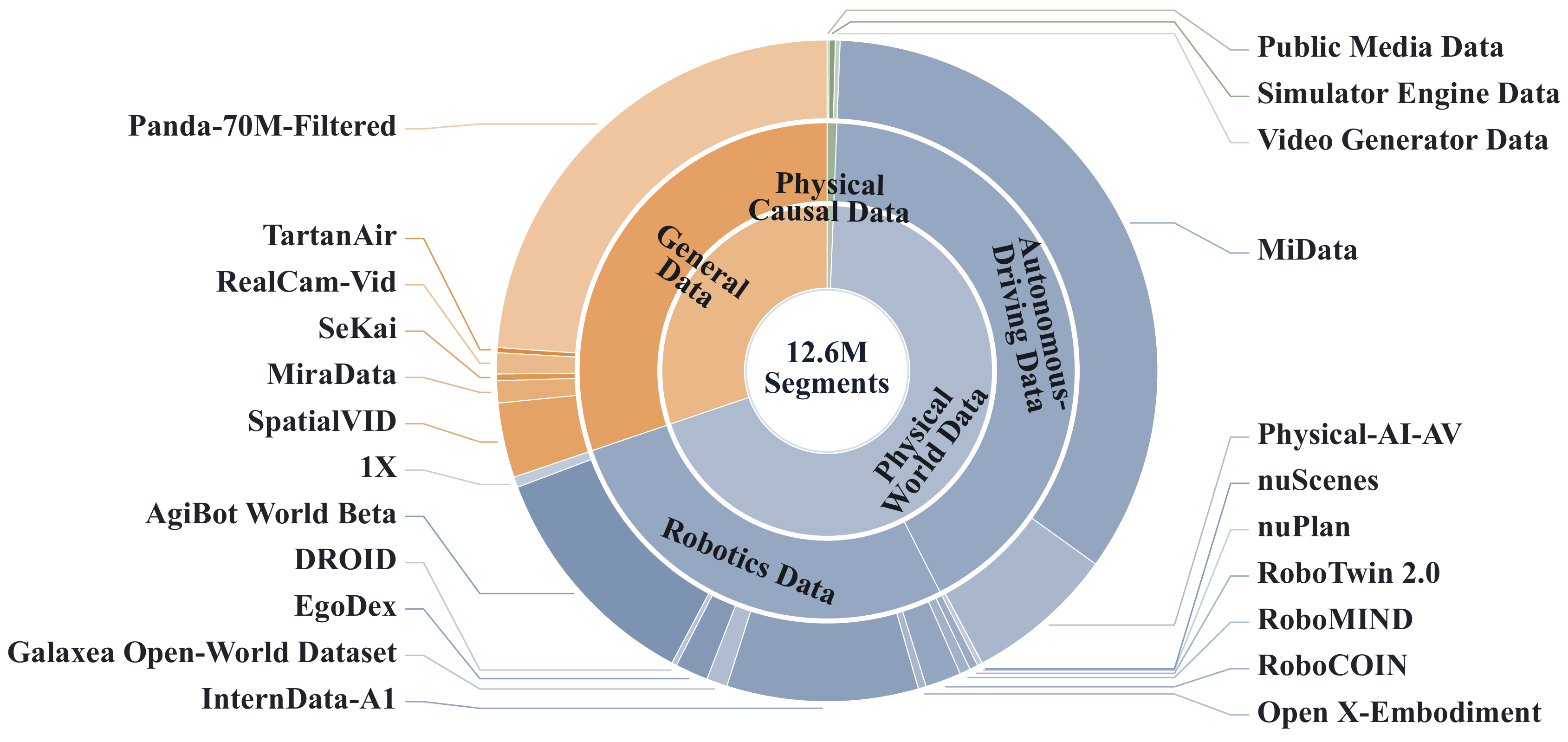}
        \caption{Dataset sample count distribution.}
        \label{fig:data_distribution_samples}
    \end{subfigure}
    \hfill
    \begin{subfigure}[b]{0.48\textwidth}
        \centering
        \includegraphics[width=1.0\linewidth]{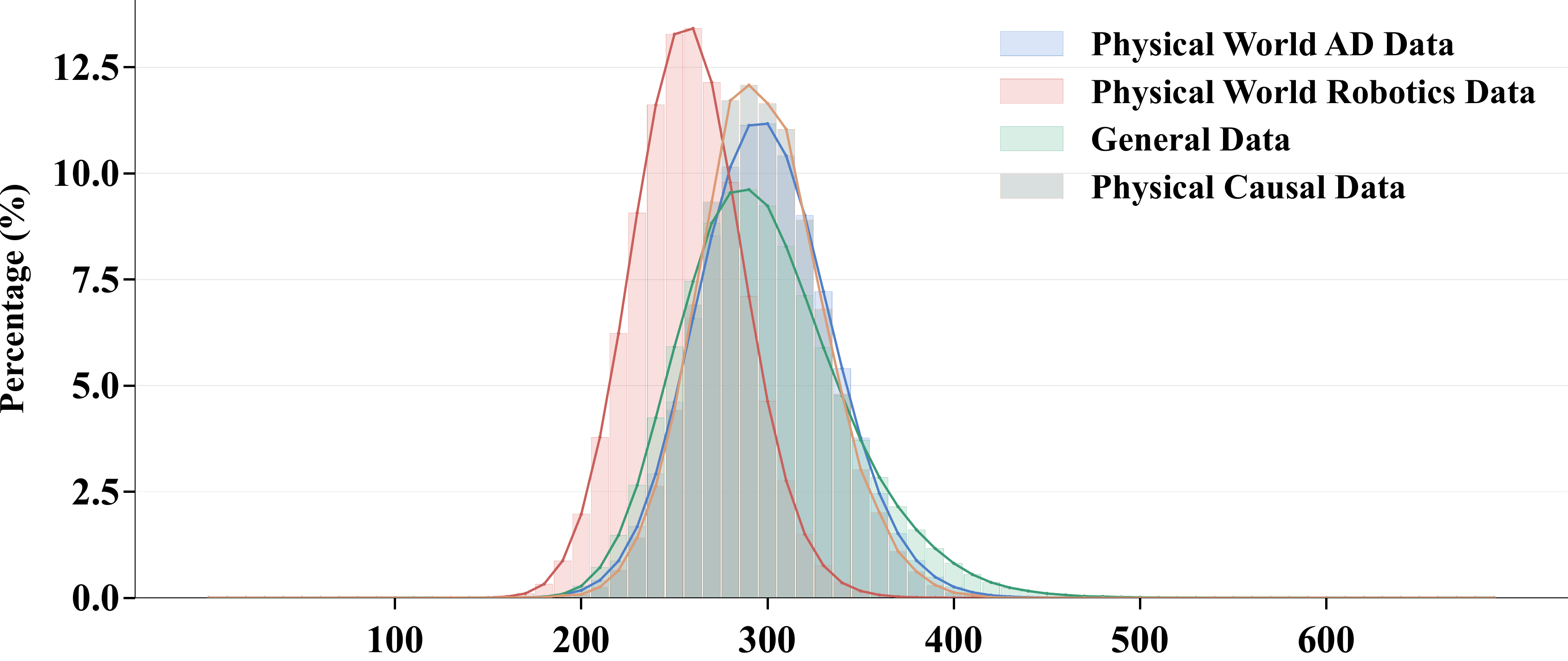}
        \caption{Caption token count distribution.}
        \label{fig:data_distribution_caption}
    \end{subfigure}
    \hfill
    \begin{subfigure}[b]{0.48\textwidth}
        \centering
        \includegraphics[width=\linewidth]{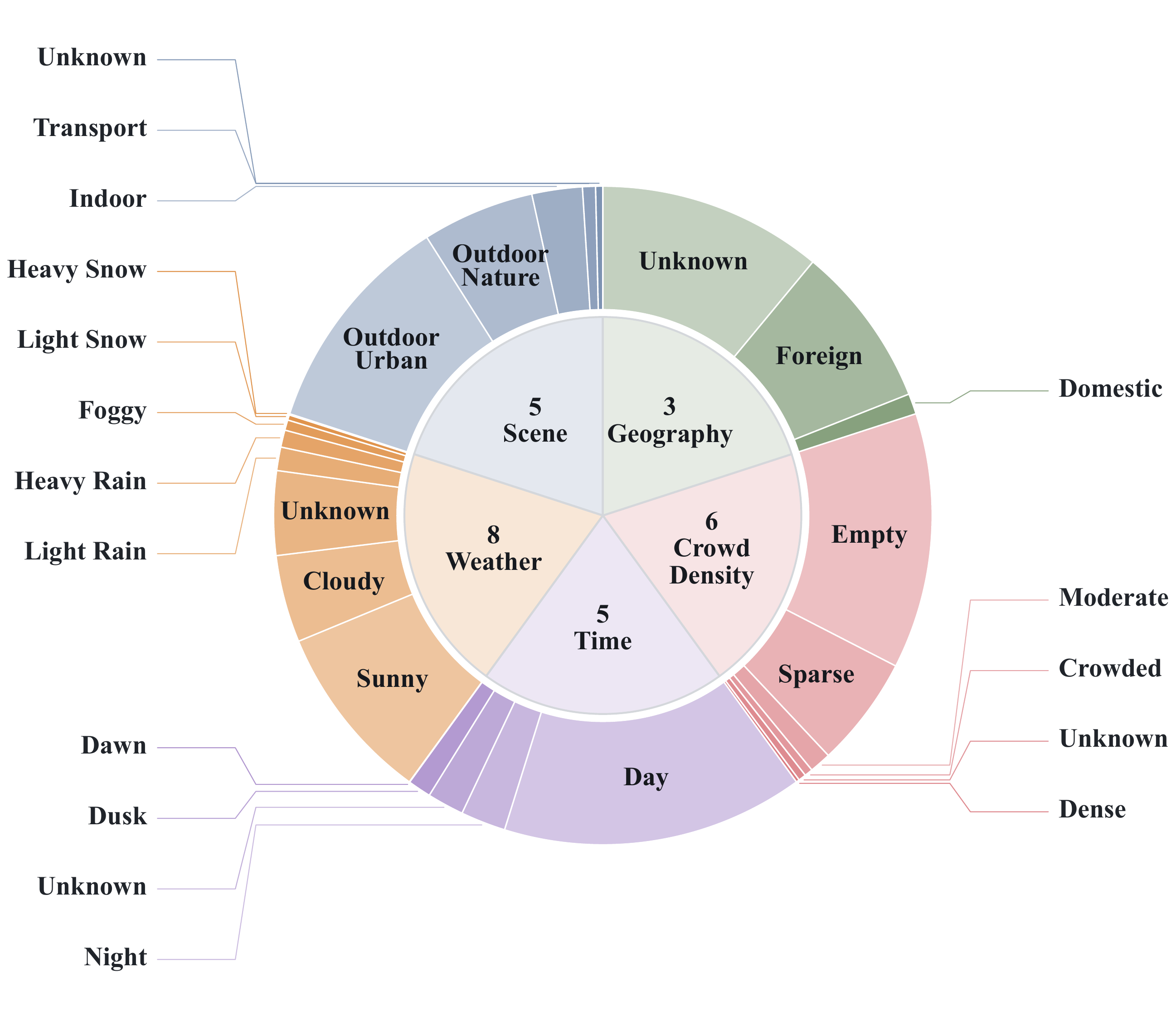}
        \caption{Scene distribution of general data.}
        \label{fig:data_distribution_general_tags}
    \end{subfigure}
    \hfill
    \begin{subfigure}[b]{0.48\textwidth}
        \centering
        \includegraphics[width=\linewidth]{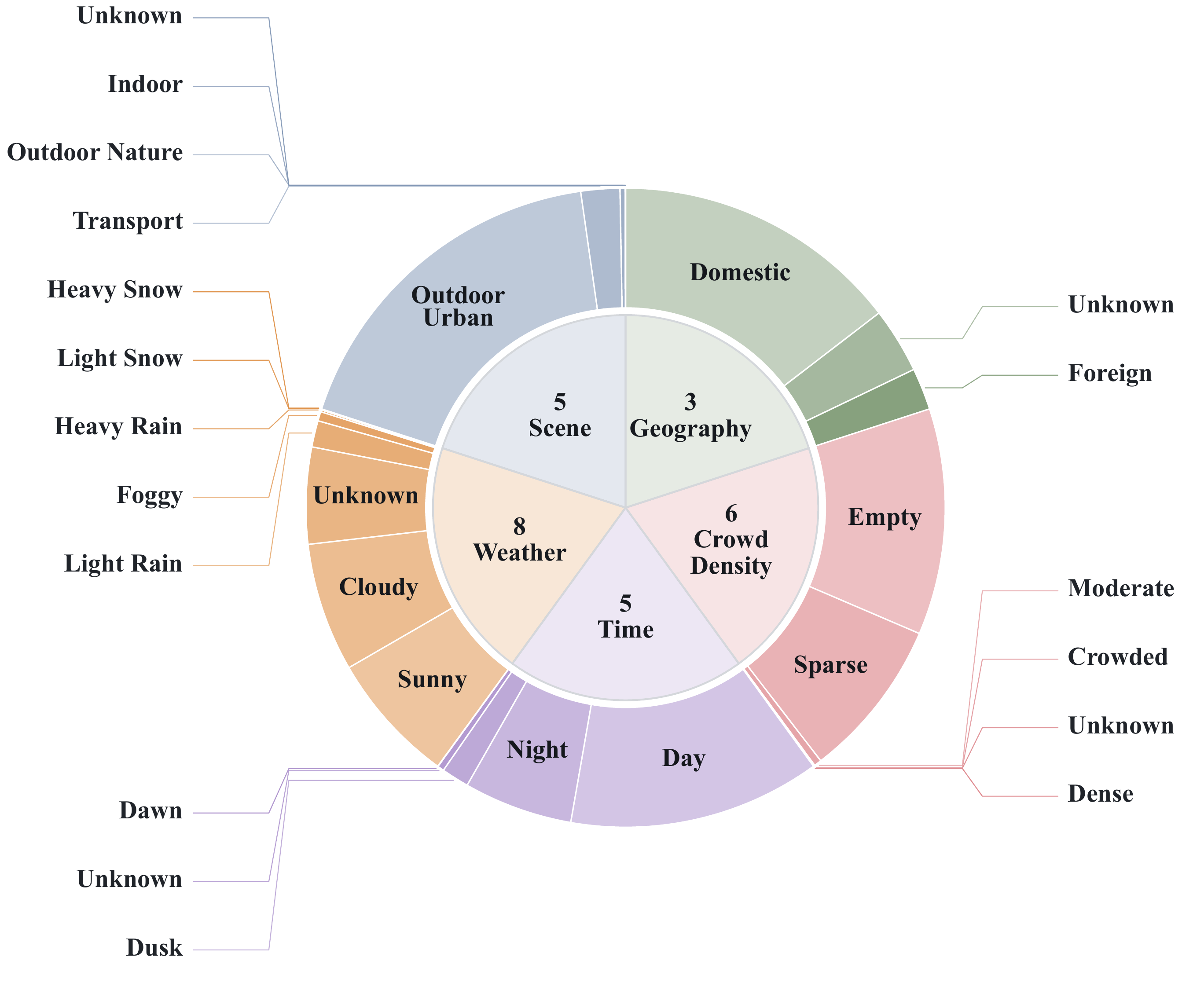}
        \caption{Scene distribution of physical world driving data.}
        \label{fig:data_distribution_drive_tags}
    \end{subfigure}
    \caption{Overview of the dataset distribution. We report the sample count distribution, caption token count distribution, and scene composition of the general and physical-world autonomous-driving data. Robotics data are omitted because they are predominantly collected in a small number of indoor settings.}
    \label{fig:data_distribution}
\end{figure}

\subsection{Dataset Distribution Analysis}


To support the effective mixing of data from different sources in subsequent training stages and guide the selection of data mixture proportions and training configurations, we further analyze the final training dataset distribution from three perspectives: sample count, caption length, and scene composition. As shown in \figref{fig:data_distribution_samples}, our training corpus contains 12.6M training clips in total, in which physical world autonomous-driving data account for the largest share, followed by a substantial amount of physical-world robotics data and general-purpose data. \figref{fig:data_distribution_caption} shows the distribution of caption token count. For most data sources, captions contain between 150 and 450 tokens, whereas robotics data tend to have shorter captions because their scenes are simpler and contain fewer objects and interactions.

For scene composition, we analyze the general and physical-world autonomous-driving subsets along five dimensions: scene type, geographic location, crowd density, time of day, and weather condition. These dimensions contain five, three, six, five, and eight categories, respectively. Each dimension includes an explicit \texttt{unknown} category for cases where the corresponding attribute cannot be reliably inferred from the caption. As shown in \figref{fig:data_distribution_general_tags} and \figref{fig:data_distribution_drive_tags}, physical driving data are dominated by outdoor-urban scenes (88.8\%), followed by transport scenes (9.8\%), whereas general data provide broader scene coverage, including outdoor-urban (55.1\%), outdoor-nature (27.6\%), and indoor scenes (12.4\%). Sunny or cloudy conditions account for 65.9\% of physical driving samples and 65.3\% of general samples. Daytime footage is more prevalent in general data (74.1\%), while nighttime footage represents a larger fraction of physical driving data (27.5\%). In terms of crowd density, scenes labeled as empty or sparse, indicating no or few people, account for 97.7\% of physical driving samples and 90.0\% of general samples. The geographic distributions also differ: physical driving data are primarily domestic (72.8\%), whereas general data contain 39.9\% foreign and 5.0\% domestic samples, with the remaining 55.0\% assigned to \texttt{unknown}.
Physical-world robotics data and physical causal data are omitted from the scene-composition plots. Robotics data are predominantly collected in a limited number of indoor settings and therefore exhibit little scene diversity, whereas physical causal data are generated from carefully designed and controlled scene configurations whose distributions are intentionally specified rather than naturally sampled.


\section{Causal World Model}
\label{sec:cwm}
To better serve Physical AI applications such as closed-loop evaluation and closed-loop training, a world model should satisfy three requirements: \textit{interactivity, real-time performance, and controllability}.
Interactivity means the model can be rolled out step-by-step in an autoregressive manner, generating the next short-horizon future conditioned on the current state and control inputs.
Real-time performance means each generation step must finish within the time budget of the simulated step, so the closed-loop system is not blocked by model inference.
Controllability means the generated futures should faithfully follow the provided control conditions (e.g., actions, world states, or camera poses), while remaining realistic and consistent with the data distribution.
These three requirements jointly impose a major challenge for model design: improving quality and controllability often calls for larger models or more denoising steps, which increases latency and can undermine both real-time performance and interactivity.
To address this challenge, we make dedicated efforts on both (i) the model architecture and condition injection, and (ii) a multi-stage progressive training pipeline with distillation-based acceleration.

\begin{figure}[t]
  \centering
  \includegraphics[width=\linewidth]{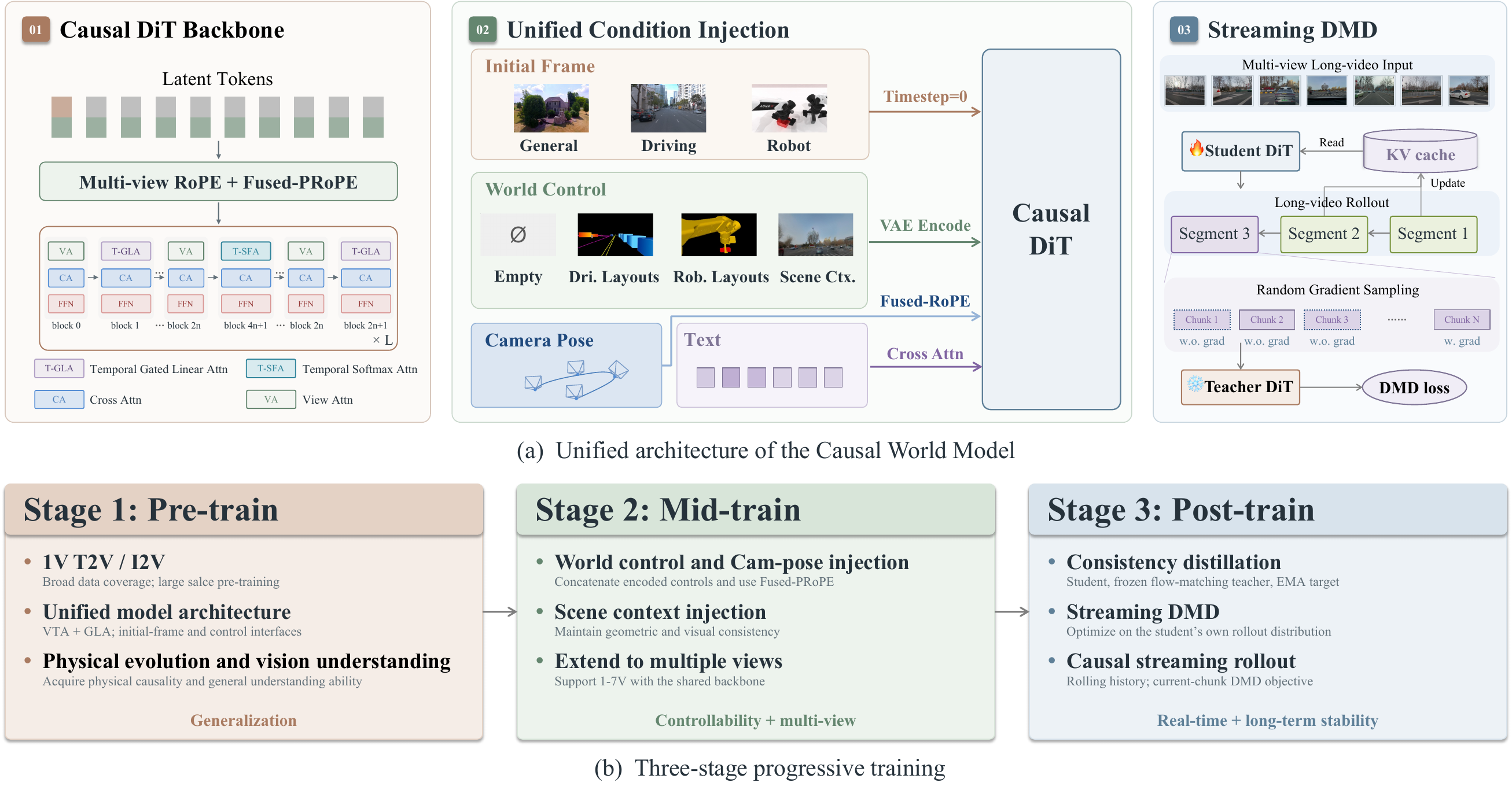}
  \caption{Architecture and training pipeline overview. The Causal World Model uses VTA and GLA to retain an efficient DiT backbone under multi-view settings. To handle diverse conditions, we adopt a unified condition injection design, enabling training across multiple tasks and datasets. During post-training, we introduce long-video rollout to improve long-horizon generation stability and Random Gradient Sampling to reduce memory usage. Overall, the model is trained in three stages to achieve efficient and controllable streaming generation.}
  \label{fig:cwm-arch-training}
\end{figure}

In this section, we first introduce the overall architecture of our Causal World Model (\secref{sec:cwm-arch}) as shown in \figref{fig:cwm-arch-training},
followed by the unified architecture adopted during pre-training (\secref{sec:cwm-pretrain}),
the condition injection strategy used in mid-training (\secref{sec:cwm-midtrain}),
and distillation in post-training (\secref{sec:cwm-posttrain}).

\subsection{Architecture}
\label{sec:cwm-arch}
Our Causal World Model adopts the architecture, weights, and VAE encoder of Wan2.2-5B~\cite{wan}, while introducing a series of modifications to the model architecture and multimodal condition injection.
Formally, we denote the model inputs as the text description $c$, the initial-frame image $I_0$, the world control $w_i$, and the camera view pose $p_i$. With these inputs, the Causal World Model is defined as
\begin{equation}
    u_{\theta}\left(x_{1:T}\mid
    w_{1:T},
    p_{1:T},
    I_0,
    c
    \right)
    =
    \prod_{i=1}^{T}
    u_{\theta}
    \left(
    x_i
    \mid
    x_{1:i-1},
    w_{1:i},
    p_{1:i},
    I_0,
    c
    \right).
\end{equation}

To unify modeling across application settings with different camera counts and world-control types, we introduce a series of improvements to the base model. In DiT, we employ VTA and GLA to model multi-view inputs more effectively and improve efficiency, and we use Fused RPoPE to encode camera-view pose control. To maintain stronger geometric and appearance consistency in closed-loop evaluation and closed-loop training, we further provide scene context with the new WorldRec2.0 algorithm.

\subsection{Pre-train}
\label{sec:cwm-pretrain}
During pre-training, we aim to endow the model with generalizable video generation capabilities. We attribute such generalization primarily to large-scale training data; therefore, to maximize data utilization and coverage, we restrict this stage to single-view T2V and I2V tasks.
However, the subsequent Mid-train and Post-train stages involve heterogeneous conditional inputs and multi-view generation. To keep a unified architecture throughout all stages, we incorporate the required architectural changes already during pre-training, including the world control injection, the VTA module and the GLA module.

\paragraph{World Control Injection.}
Our Causal World Model adopts a lightweight and general approach to inject world control into the model.
Since the world control $w_i$ has the same dimensionality as the target $x_i$, we directly encode $w_i$ into the latent space using the same VAE and concatenate it with the target $x_i$ before feeding them into the model.
We re-initialize the projection layer to ensure that the token dimension fed into the DiT Transformer~\cite{dit} matches that of the base model, thereby maximizing the inheritance of the capabilities from base model.
Compared with approaches such as ControlNet~\cite{controlnet} or Cross-Attention, this design is more lightweight and general, introduces no additional computational cost, and does not alter the architecture or weight parameters of the base model.

During pre-training, the vast majority of training samples do not have corresponding $w_i$. For such samples, we apply zero padding to $w_i$, thereby ensuring that the world control input is handled consistently across the pre-training, Mid-train, and Post-train stages.
For the initial-frame image $I_0$, following Wan2.2, we replace the first frame of the noisy latent with $I_0$ and set the corresponding timestamp to $0$.

\paragraph{View Temporal Attention.}
VTA is designed to address the computational overhead introduced by an increasing number of views. It reduces the growth of Attention computation while maintaining generation quality and cross-view consistency, thereby preserving inference efficiency.
Specifically, our Causal World Model consists of $L$ DiT Blocks, each of which contains a Cross-Attention module and a Self-Attention module.
For a sample with $N$ views and a temporal length of $T$, 
the original Self-Attention flattens the view, temporal, and spatial dimensions into a single sequence, and thus has an attention complexity of $O\left(N^2T^2\right)$.
To reduce the computational cost of multi-view modeling, VTA factorizes Self-Attention along the temporal and view dimensions and organizes the DiT Blocks into alternating odd and even layers.
Odd-numbered layers perform only Temporal Attention, with $N$ folded into the batch dimension.
Even-numbered layers perform only View Attention, with $T$ folded into the batch dimension.
After alternately stacking $L$ layers, the core complexity, ignoring the batch size and the spatial dimensions, becomes $O\left(TN(N+T)\right)$, which maintains high inference efficiency in both single-view and multi-view settings.

Compared with other approaches, such as OmniDreams~\cite{OmniDreams}, our VTA does not introduce additional Attention modules and computes.
During pre-training, to ensure consistency with subsequent multi-view training, the View Attention degenerates into intra-frame Self-Attention. Our experiments show that this design does not lead to a significant performance degradation.
With this design, the model establishes a unified mechanism for multi-view correlation modeling, and our DiT Blocks require no modification when extending the model to multi-view tasks, enabling multi-view models to inherit the generalization capabilities learned during pre-training.

\paragraph{Gated Linear Attention.}
VTA removes the quadratic dependence on the number of views, but Temporal Attention remains quadratic in the temporal length $T$. Let $P = hw$ be the tokens per view per frame. Each Temporal Attention layer attends over $TP$ tokens, costing $O\!\left(N (TP)^{2} C\right)$, and its KV cache grows linearly with rollout length. By contrast, View Attention uses a sequence length $NP$ that is independent of $T$ and needs no cross-timestep cache. We therefore apply gated linear attention only to Temporal Attention.

We use the gated bidirectional linear attention from SANA-Video~2.0~\cite{sanavideo2}, which compresses the sequence into a fixed $d \times d$ state per head ($d$ is the head dimension). Let $q_{j}^{r}, k_{n}^{r}$ be RoPE-rotated~\cite{su2024roformer} queries/keys and $\phi(\cdot)=\operatorname{ReLU}(\cdot)$; each head computes
\begin{equation}
S = \sum_{n} v_{n} \big( \beta_{n} k_{n}^{r} \big)^{\!\top},
\qquad
o_{j} = W_{o}\!\left[
  \operatorname{RMSNorm}\!\left(
    \frac{S\, q_{j}^{r}}{\sum_{n} \phi(k_{n})^{\!\top} \phi(q_{j}) + \epsilon}
  \right) \odot \sigma(g_{j})
\right],
\label{eq:gla}
\end{equation}
where $\beta$ gates writes to the state and $g$ gates the output~\cite{qiu2025gated}. Because $S$ is a permutation-invariant sum, the operator is bidirectional, matching our bidirectional pre-training. For causal training, we restrict the sum to the chunk prefix (bidirectional within a chunk, causal across chunks), reusing the same expression and parameters, so it transfers to autoregressive rollout without dropping weights. Both stages are warm-started from the pretrained base checkpoint; the added gate projections are zero-initialised with a positive bias so that $\beta$ and $g$ start near identity.

GLA reduces both compute and memory in the Temporal Attention branch. We do not linearize every Temporal Attention layer; instead, we keep periodic gated-softmax anchors~\cite{sanavideo2,qwen3next,kimilinear}. We use a $4{:}1$ schedule, which matches the full-softmax baseline in our experiments while leaving fewer quadratic layers. With $L$ blocks, anchors are placed at block $4n+1$ (see~\figref{fig:cwm-arch-training}), so shallow, intermediate, and deep features each receive at least one full mixing pass. During streaming inference, each temporal GLA layer stores history in a length-independent state, reducing the per-layer temporal cache by over $97\%$ relative to a softmax KV cache and making the per-step cost independent of generated length.

\subsection{Mid-train}
\label{sec:cwm-midtrain}
After pre-training, the Causal World Model already exhibits general generation capability, but it still needs to be further extended to support conditional generation across diverse physical AI applications.
The mid-train stage has two concurrent goals: (i) enabling control conditions such as world control $w_i$ and camera view pose $p_i$ to reliably steer the generated state, and (ii) extending the model to generate multi-view state sequences.

For world control, we concatenate it into the latent-space channel reserved during pre-training.
For camera view-pose control and multi-view extension, we achieve this by modifying the original 3D RoPE~\cite{wan,su2024roformer} which also introduces no additional model parameters or computational overhead.
Overall, the model architecture remains consistent with the pre-training stage, so that it can inherit the generalization ability and physical reasoning learned during pre-training. 

\paragraph{Multi-view RoPE.}
We extend the original 3D RoPE to support positional encoding in the multi-view setting.
The original 3D RoPE partitions frequencies along the temporal, height, and width dimensions.
For multi-view scenarios, we allocate non-overlapping coordinates to different views along the width axis. Concretely, for a sample with $N$ views and temporal length $T$, we horizontally concatenate all views at the same time step, yielding
$x \in \mathbb{R}^{T \times h \times (Nw) \times d}$,
where $w$ is the per-view width.
A token at local coordinate $(i,u,v)$ in view $n$ is assigned the absolute coordinate $(i,h,nW+w)$.

All views share the same temporal and height origins, while the width origin is shifted by $nw$.
In implementation, Temporal Attention first splits the concatenated sequence into $N$ single-view sequences of shape $T\times h\times w$, and then applies the original 3D RoPE with the corresponding width offset.
As a result, relative positions within each view are identical to the single-view case, while different views lie on distinct horizontal position bands.

\paragraph{Fused PRoPE.}
Further, we condition our Causal World Model on camera view-pose with fused projective rotary positional embeddings (Fused-PRoPE), following SolarWM~\cite{solarwm} and MosaicMem~\cite{Mosaicmem}.
Each multi-view clip is converted into a geometric condition aligned with the latent video tokens.
Per-frame, per-view poses are expressed relative to a shared origin, namely the first frame of a center camera together with its corresponding intrinsics. Within each DiT block, Multi-view RoPE represents temporal and intra-view spatial positions, and camera view-pose control $p_i$ determines the projective matrix.

\begin{equation}
P_i = K_f E,
\end{equation}
where \(E\) denotes the relative extrinsics and  focal-only
intrinsics \(K_f\).
We apply \(P_i^{\top}\) to queries and \(P_i^{-1}\) to keys and values, run attention, and map the output back with \(P_i\) before the output projection.
The same injection is used in both branches of view-temporal attention: even blocks attend over time within a camera, odd blocks attend across views at a fixed time.
Camera motion is therefore introduced through the attention computation itself, rather than by appending an unrelated condition token or a separate camera branch.


\subsection{Post-train}
\label{sec:cwm-posttrain}
Although the model acquires controllable generation capabilities through pre-training and mid-training, it still cannot meet the requirements of real-time streaming inference.
Moreover, during inference, the model suffers from exposure bias. We therefore further distill the model to improve both inference efficiency and long-term generation stability.

\paragraph{Consistency Distillation.}
We first perform few-step distillation using a consistency-model objective~\cite{consistencymodels,causalforcing++}. Specifically, we treat the autoregressive model trained in mid-training as the teacher. The student is initialized with the same weights and set to be trainable. During consistency distillation, given the state $x_i^t$, where $t$ is diffusion timestep, the teacher predicts the next state $\tilde{x}_i^{t-\delta t}$ along the same probability-flow ODE (PF-ODE) trajectory. The student network $G_\theta$ also predicts from $x_i^t$ and is trained to match the prediction of the stop-gradient EMA target network $G_{\theta^-}$ evaluated at $\tilde{x}_i^{t-\delta t}$:
\begin{equation}
\mathcal{L}=\mathbb{E}\big[d\big(G_\theta(x_i^t),\, G_{\theta^-}(\tilde{x}_i^{t-\delta t})\big)\big],
\end{equation}
where $d(\cdot,\cdot)$ denotes distance under a predefined norm and $G_{\theta^-}$ is the EMA of $G_\theta$ with gradients detached. 
With this objective, we distill the multi-step teacher into a few-step student while retaining generation quality.

\paragraph{Streaming DMD.}
After the above consistency distillation, our model still suffers from exposure bias during long-video inference. A common remedy is to apply DMD~\cite{dmd} with self-forcing-style training. However, standard self-forcing is only trained with a short horizon, so the model cannot fully capture long-range temporal dependencies, which still leads to suboptimal quality for long-video generation. Following LongLive~\cite{longlive}, we extend self-forcing~\cite{selfforcing} to a streaming DMD training scheme, as illustrated in \figref{fig:cwm-arch-training}(a). Specifically, we train on long videos and split each sample into multiple segments, where each segment length matches the training length of the bidirectional teacher. During training, the student continuously rolls out the same sample across multiple consecutive steps; the rollout results are then fed to the teacher model and a fake model to compute the DMD loss.

Since a video may contain up to seven views, naively rolling out all segments and backpropagating the DMD loss can easily trigger GPU out-of-memory. To address this, we introduce a Random Gradient Sampling strategy. Each segment is composed of multiple chunks. Instead of recording gradients for every chunk as in prior implementations, we only keep gradients for the last chunk and randomly select a small number of earlier chunks to record gradients. In practice, this achieves training behavior comparable to full-chunk backpropagation while substantially reducing memory consumption.

\subsection{WorldRec 2.0}
\label{sec:worldrec2}

\begin{figure}[t]
  \centering
  \includegraphics[width=\linewidth]{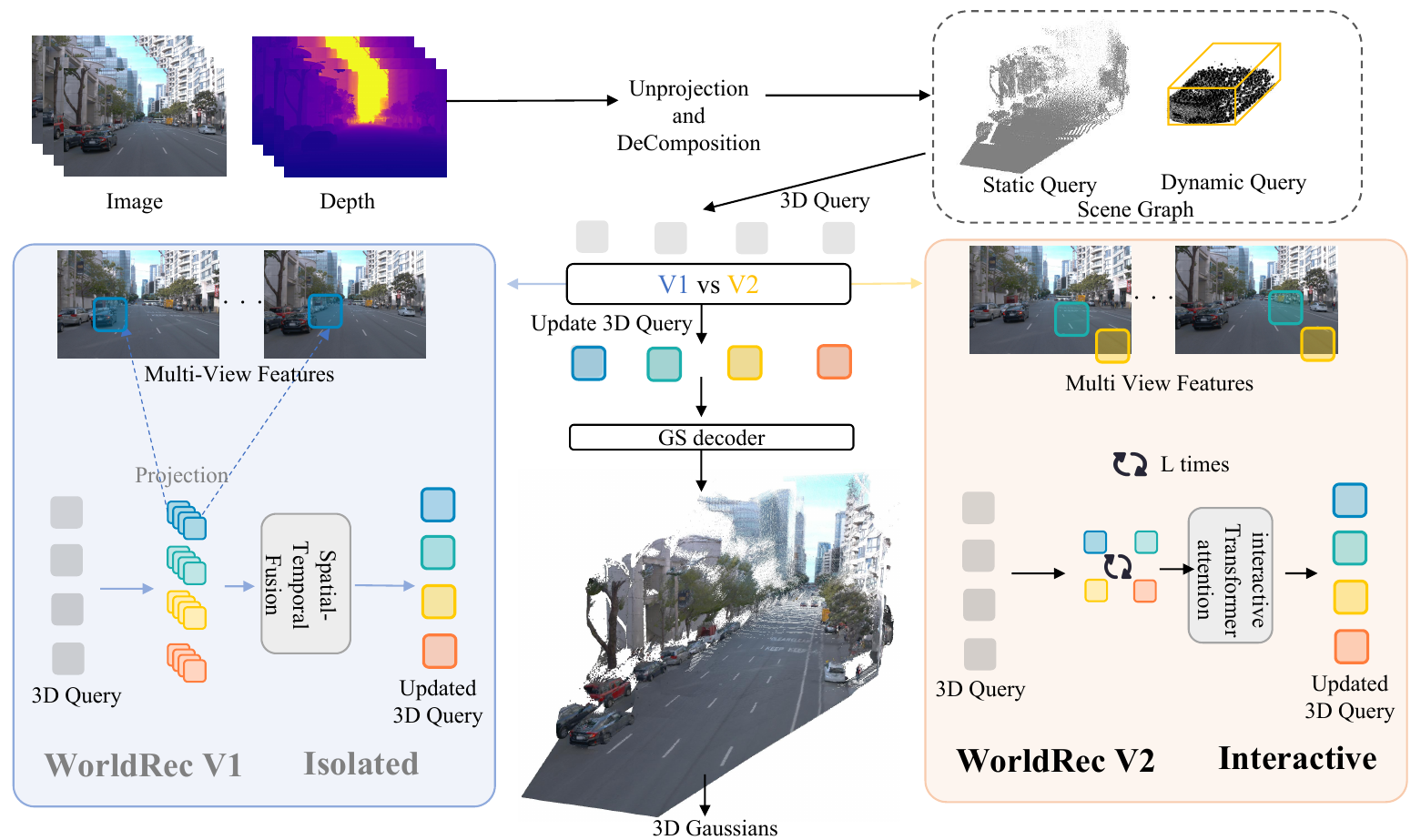}
  \caption{\textbf{WorldRec 2.0: point-aligned interaction and explicit 3D scene memory.}
Multi-camera and multi-temporal observations are aligned in 3D, exchanged
through static-scene and dynamic-instance registers, decoded into static and
dynamic Gaussians, and rendered under a target camera pose and
time.}
  \label{fig:WorldRec2_MAIN}
\end{figure}

To enhance consistency and stability in long-horizon causal inference, we introduce an explicit scene context grounded in real observations. Concretely, we propose WorldRec 2.0 to construct a persistent 3D scene state and use it as a geometry-aware prior for generation. Given multi-camera, multi-temporal observations, WorldRec 2.0 reconstructs a scene state composed of static and dynamic 3D Gaussians in a single forward pass, and renders it under a target camera and timestamp. The rendered prior is then fed into the generator through the world-control channels introduced in \secref{sec:cwm-midtrain}, serving as a stable 3D scene memory for causal generation. In this way, WorldRec 2.0 anchors geometry, appearance, and cross-view correspondence in observed regions, while the Causal World Model generates more realistic scenes and completes fine-grained appearance details as well as motion details beyond the observations.

\paragraph{From Point-Aligned Representation to Point-Aligned Interaction.}
WorldRec 1.0~\cite{XiaomiAutoWorldModel1} follows the point-aligned design of PointForward 1.0~\cite{chi2026pointforward}. It lifts multi-view images with estimated depth into a shared 3D space, and aggregates observations from the same spatial location into a set of sparse 3D queries. Each query is initialized from its 3D position, color, image feature, and Pl\"ucker ray. Static queries stay in world coordinates. For each dynamic instance $k$, observed points are mapped into a canonical coordinate system using the instance pose $T_k(\tau)$:
\begin{equation}
    \mathbf{q}=\phi_{\mathrm{init}}(\mathbf{x},\mathbf{r},\mathbf{c},\mathbf{f}),
    \qquad
    \mathbf{x}_{\mathrm{can}}=T_k(\tau)^{-1}\mathbf{x}_w.
\end{equation}
This point-aligned representation helps reduce duplicate points and the ghosting artifacts introduced by frame-wise Gaussian prediction. However, fusion mainly happens within each query, and related queries lack an efficient communication path. In contrast, applying full global attention to tens of thousands of queries would be prohibitively expensive in both compute and memory.

\paragraph{Scene-Structured Register Interaction.}
To enable efficient communication, WorldRec 2.0 adopts the register-based interaction design of PointForward 2.0 ~\cite{wang2026pointforward}. We introduce a small set of learnable registers as a shared communication bottleneck among point queries. Each interaction block consists of three stages---compress, communicate, and broadcast:
\begin{equation}
    \hat R=A(R,Q),\qquad
    \tilde R=S(\hat R),\qquad
    \tilde Q=A(Q,\tilde R),
\end{equation}
where $A$ denotes query--register cross-attention and $S$ denotes register self-attention. Static scene registers propagate background information across views and timestamps. Dynamic instance registers aggregate observations of the same object over time in its canonical space, while separating unrelated instances. On the query side, we additionally employ 4D RoPE built from the aligned 3D coordinates and the original timestamp, so that temporal provenance is preserved after spatial alignment. Overall, this scene-structured interaction supplies the Gaussian decoder with richer geometric and appearance context without resorting to point-level global attention.

\paragraph{Gaussian Scene Memory and Generative Interface.}
The updated queries are directly decoded into RGB Gaussians parameterized by position, color, scale, rotation, and opacity. Static Gaussians are represented in world coordinates. Dynamic Gaussians are predicted in canonical coordinates and then transformed back to world space using the instance pose at the target timestamp. We train the reconstruction with RGB, depth, and perceptual objectives.

Let $\mathcal{G}_s$ denote the static Gaussian set and $\mathcal{G}_d^k$ denote the canonical Gaussian set of dynamic instance $k$. Given a target camera pose $p_i$ and the instance poses $T_k(i)$, we render a geometry-aware prior as
\begin{equation}
    r_i=\mathcal{R}\!\left(
    \mathcal{G}_s\cup
    \bigcup_{k=1}^{K}T_k(i)\mathcal{G}_d^k,\,
    p_i\right),
    \qquad
    \tilde w_i=[w_i;r_i],
\end{equation}
\noindent where $\mathcal{R}$ is a Gaussian rasterizer. The augmented world control $\tilde w_i$ is injected into the DiT via the VAE encoding and channel-concatenation mechanism in \secref{sec:cwm-midtrain}, without introducing a separate conditioning branch. In effect, WorldRec 2.0 writes observation-supported content into a persistent 3D scene memory, providing a stable geometric anchor across time and views during streaming rollout. We report detailed reconstruction results in \secref{sec:worldrec2_experiments}.

\section{Training and Inference}
\label{sec:train-infer}

\subsection{Multi Stage Training}
Our Causal world model training process is mainly divided into three stages. The training data
and data mixture differ across stages, but we keep the model architecture consistent throughout,
allowing the weights from the previous stage to be reused as the initialization for the next stage.
The specific data usage is summarized in \tabref{tab:train-curriculum}.

\begin{table}[t]
    \centering
    \caption{Multi-stage training curriculum. In General Data, \emph{Full} denotes the complete dataset, and HQ denotes the high-quality subset selected using the scoring scheme in Chapter~2. In Physical Causal Data, \emph{Pub.\ Media} refers to counterfactual video data retrieved from public datasets and open-source materials (e.g., vehicle-collision clips).}
    \label{tab:train-curriculum}
    \small
    \setlength{\tabcolsep}{5.5pt}
    \renewcommand{\arraystretch}{1.15}
    \resizebox{\linewidth}{!}{%
    \begin{tabular}{lccccccccc}
      \toprule
      \multirow{2}{*}{Data \& Stage} & \multirow{2}{*}{Res.} & \multirow{2}{*}{Views} & \multicolumn{2}{c}{General Data} & \multicolumn{2}{c}{Physical World Data} & \multicolumn{3}{c}{Physical Causal Data} \\
      \cmidrule(lr){4-5} \cmidrule(lr){6-7} \cmidrule(lr){8-10}
      &  &  & Full & HQ & Driving & Robot & Engine & \begin{tabular}[c]{@{}c@{}}LVM\end{tabular} & \begin{tabular}[c]{@{}c@{}}Pub. Media\end{tabular} \\
      \midrule
      Data samples & -- & -- & 3.6M & 0.5M & 5.4M & 3.6M & 37K & 29K & 12K \\
      Pre-train     & 480p & 1V & \checkmark & \checkmark & \checkmark & \checkmark & \checkmark & \checkmark & \checkmark \\
      Mid-train   & 480p,540p & 1-7V & -- & \checkmark & \checkmark & \checkmark & \checkmark & \checkmark & \checkmark \\
      Post-train & 480p,540p & 1-7V & --         & --         & \checkmark & \checkmark & -- & -- & -- \\
      \bottomrule
    \end{tabular}%
    }
  \end{table}

\paragraph{Stage~I: Pre-training.}
In Stage~I, our model architecture is identical to that described in \secref{sec:cwm-pretrain}.
Using this architecture, we train two models separately: a bidirectional model and an autoregressive model.
Both models are trained with the flow-matching loss.
Given a sample $x_{1:T}$, we denote its corresponding clean latent as $x^{0}$ and random noise as $x^{1} \sim \mathcal{N}(0,I)$ (we omit subscripts for simplicity).
We sample a timestep $t \in [0,1]$ from a logit-normal distribution and define the intermediate latent $x^{t}$ and its corresponding ground-truth velocity $v^{t}$ as
\begin{equation}
x^{t} = (1-t) x^{0} + t x^{1}, \qquad v^{t} = x^{1} - x^{0}.
\end{equation}

We feed $x^{t}$ into the model to predict the velocity
$u_{\theta}\left(x^{t} \mid w, p, I_0, c \right)$.
and the loss is computed using MSE.
For the bidirectional model, all chunks $x_i^t$, $i \in \{1,\ldots,T\}$, use the same timestep $t$,
and full attention is applied between all chunks.
For the unidirectional model, we use Diffusion Forcing for training and assign a different timestep $t_i$ to each chunk $x_i^{t_i}$.
We apply a causal attention mask, so that $x_i^{t_i}$ can attend only to the historical chunks $x_{1:i-1}$. 
During training, we use a learning rate of $1\times 10^{-4}$ and the AdamW optimizer with
$\beta_1=0.9$, $\beta_2=0.95$, and a weight decay of $5\times 10^{-2}$.
To facilitate optimization, we first train the bidirectional model and use its learned parameters to initialize the Diffusion Forcing~\cite{diffusionforcing} model. 
We group every two latent frames into one chunk (i.e., \texttt{chunksize}=2).

\paragraph{Stage~II: Mid-train.}
In Stage~II, our goal is to endow the model with controllability and the ability to generate multi-view states, while preserving its general generation capability. We therefore adopt a progressive, multi-step training curriculum. First, we mix single-view data with multi-view data and train with T2V and I2V tasks, where the single-view data comes from the high-quality subset of General Data and Physical World Data, and the multi-view data comes from driving Data and robot Data. Next, we introduce camera view-pose conditioning with a certain probability, resulting in a hybrid training objective over T2V, I2V, and CI2V. Finally, we incorporate world-control conditions and train with T2V, I2V, CI2V, WI2V, and WCI2V. We design this curriculum because CI2V is more challenging and has more data than WCI2V; introducing the harder task earlier helps the model learn complex conditional generation. In this stage, we initialize the bidirectional model and the autoregressive model separately from their Stage~I counterparts.

\paragraph{Stage~III: Post-train.}
In Stage~III, we optimize the model for downstream deployment by post-training on a small, carefully curated subset from the high-quality portions of driving Data and robot Data. We explicitly balance this subset to match the desired application-specific distribution and to ensure both high quality and diversity. This is crucial because CD and Streaming DMD are prone to mode collapse, which can lead to a substantial drop in sample diversity; by finely adjusting the training samples, we encourage the model to recover both diversity and fidelity. 
Concretely, we run post-training separately for different application scenarios. Before applying CD and Streaming DMD, we further fine-tune the bidirectional model and the autoregressive model on the corresponding scenario-specific high-quality subset, using a learning rate of $2\times 10^{-2}$ for roughly one epoch. We then perform CD and Streaming DMD to obtain scenario-specific weights (e.g., autonomous-driving 7V and embodied 3V). Meanwhile, we reduce the training objective to a fixed task (e.g., WCI2V) so that only a small number of optimization steps is needed to achieve strong downstream performance. 
During the Streaming DMD training stage, we roll out the student model for two steps and distill a model that performs \texttt{two-step} denoising with \texttt{chunksize}=2.

\subsection{Inference.}
\label{sec:train-infer-infer}
In closed-loop evaluation and training, inference throughput and latency are critical. We speed up inference along three dimensions: world-control generation, VAE encoding/decoding, and DiT denoising. First, world-control signals must be rendered online based on the ego pose and dynamic scene changes. Following FlashDreams~\cite{flashdreams2026,OmniDreams}, we use Ludus to accelerate rendering.
Second, for VAE encoding and decoding, we combine \texttt{torch.compile}~\cite{pytorch2} with CUDA Graphs~\cite{guide2020cuda} to reduce kernel-launch overhead and improve GPU utilization.
Third, for DiT denoising, we allocate a fixed-size KV cache for the local-attention window. To keep training and inference consistent, we also use a fixed KV-cache size during post-training for self rollouts. This keeps the computation graph and tensor shapes constant at inference time, enabling CUDA-Graph acceleration for DiT. We further run inference in parallel on 8 GPUs. In the 7V setting, we shard the tokens of each view across 7 GPUs, while the remaining GPU handles VAE encoding for world control. The generated latents are then decoded on the same 7 GPUs. We implement context parallelism via ring attention~\cite{liu2024ringattention}.

\definecolor{xiamiOrange}{RGB}{255, 102, 0} 
\definecolor{xiamiBlue}{RGB}{0, 102, 204}   

\begin{table}[!htbp]
\centering
\small
\caption{\textbf{Efficiency and quality of three DiT attention designs.}
\textit{FSA} denotes native full softmax attention and flattens all the view tokens;
\textit{VTA} denotes view-temporal decoupled attention with softmax attention; and
\textit{GLA} uses the VTA with GLA in temporal attention branch.
\textit{SA/layer} reports the mean CUDA-event latency (per DiT forward pass) of a representative attention layer used by each design: full spatiotemporal softmax attention for \textit{FSA}, temporal-branch softmax attention for \textit{VTA}, and temporal-branch gated linear attention for \textit{GLA}.
\textit{DiT/chunk} is the causal DiT latency per chunk generation.
Parenthesized efficiency values are speedups relative to FSA; the parenthesized FID change compares GLA with VTA.}
\label{tab:efficiency-quality_gla}
\begin{tabular}{@{}lcccc@{}}
\toprule
Attention
& SA/layer (ms) $\downarrow$
& DiT/chunk (s) $\downarrow$
& Latent FPS $\uparrow$
& FID $\downarrow$  \\
\midrule
FSA
& 60.14
& 6.481
& 0.309
& -- \\

VTA
& 23.17 (2.60$\times$)
& 2.299 (2.82$\times$)
& 0.870 (+182\%)
& \textbf{37.24} \\

GLA(\textbf{Ours})
& \textbf{8.412} (7.15$\times$)
& \textbf{1.742} (3.72$\times$)
& \textbf{1.148} (+272\%)
& 38.08 (+2.3\%) \\
\bottomrule
\end{tabular}
\end{table}

\section{Results}
\label{sec:results}

\subsection{Video Generation Evaluation}

\tabref{tab:efficiency-quality_gla} compares efficiency and generation quality across three attention designs. All models use three-stage training; we report final distilled two-step results.
All measurements use the same 7V setting and $100$ output latent frames on a single NVIDIA H20 GPU.
VTA cuts attention-layer latency from $60.14$\,ms (FSA) to $23.17$\,ms ($2.60\times$), and GLA further cuts it to $8.412$\,ms ($2.75\times$ over VTA; $7.15\times$ over FSA).
End-to-end, causal DiT latency per 2-latent-frame chunk drops from $6.481$\,s (FSA) to $2.299$\,s (VTA) and $1.742$\,s (GLA), increasing throughput from $0.309$ to $1.148$ latent frames/s.


Unless otherwise specified, all subsequent experiments use the VTA and hybrid GLA as DiT backbone. Using this shared setting, we evaluate the model in the Mid-train stage from three complementary perspectives: general video generation, scene perception, and camera-pose control.
General video generation assesses visual quality and temporal coherence on standard benchmarks, together with the ability to simulate long-tail scenarios and physically causal events across synchronized views. 
Scene perception examines whether the model captures scene geometry and preserves object-level information through depth prediction and referring video segmentation.
Camera-pose control evaluates whether explicit geometric conditions can steer viewpoint changes while maintaining the consistency of the generated scene.

\begin{table}[!htbp]
\centering
\caption{Quality evaluation on the Autonomous Vehicle subset of PAI-Bench-G. AV-Quality denotes the average score; SC, BC, MS, AQ, IQ, OC, IS, and IB denote Subject Consistency, Background Consistency, Motion Smoothness, Aesthetic Quality, Imaging Quality, Overall Consistency, Image-to-Video Subject Consistency, and Image-to-Video Background Consistency, respectively. \textbf{Bold} and \underline{underlined} values denote the best and second-best results per column.}
\label{tab:av_generation}
\resizebox{\linewidth}{!}{%
\begin{tabular}{lccccccccc}
\toprule
Model
& AV-Quality $\uparrow$
& SC $\uparrow$
& BC $\uparrow$
& MS $\uparrow$
& AQ $\uparrow$
& IQ $\uparrow$
& OC $\uparrow$
& IS $\uparrow$
& IB $\uparrow$ \\
\midrule
Wan2.2-5B~\cite{wan}
& 76.1 & 90.4 & 93.7 & 97.1 & 47.5 & 67.4 & 21.8 & 95.4 & 95.8 \\
Wan2.1-14B~\cite{wan}
& 77.2 & 91.7 & 94.3 & 98.4 & 48.1 & 68.8 & 21.7 & 97.0 & 97.3 \\
Wan2.2-14B~\cite{wan}
& 77.0 & 91.4 & 94.1 & 98.6 & 47.3 & 68.1 & 21.9 & 97.2 & 97.5 \\
DreamXWorld~\cite{dreamxworld}
& 76.1 & \underline{91.9} & 93.7 & 97.3 & 47.5 & 64.4 & 21.7 & 95.7 & 96.7 \\
HunyuanVideo1.5~\cite{hunyunavideo}
& \underline{77.4} & 91.5 & 92.5 & \underline{99.3} & \textbf{48.8} & \underline{69.0} & \textbf{22.3} & \underline{97.9} & \underline{98.1} \\
LingBotWorld~\cite{lingbotworld}
& 77.3 & 91.2 & \underline{94.4} & 98.8 & \underline{48.3} & \textbf{69.1} & 21.8 & 97.3 & 97.8 \\
\textbf{Ours}
& \textbf{78.0} & \textbf{96.2} & \textbf{96.5} & \textbf{99.5} & 46.6 & 64.5 & \underline{22.2} & \textbf{99.0} & \textbf{99.2} \\
\bottomrule
\end{tabular}%
}
\end{table}

\begin{table}[!htbp]
\centering
\caption{Quality evaluation on the Robot subset of PAI-Bench-G using the same metrics as \tabref{tab:av_generation}. Robot-Quality denotes the average score. 
}
\label{tab:robot_generation}
\resizebox{\linewidth}{!}{%
\begin{tabular}{lccccccccc}
\toprule
Model
& Robot-Quality $\uparrow$
& SC $\uparrow$
& BC $\uparrow$
& MS $\uparrow$
& AQ $\uparrow$
& IQ $\uparrow$
& OC $\uparrow$
& IS $\uparrow$
& IB $\uparrow$ \\
\midrule
Wan2.2-5B~\cite{wan}
& 74.3 & 89.0 & 91.0 & 98.0 & 46.1 & 67.8 & 19.2 & 90.3 & 92.8 \\
Wan2.1-14B~\cite{wan}
& 75.2 & 87.9 & 90.3 & 98.5 & 48.8 & \textbf{71.7} & 19.0 & 91.5 & 93.7 \\
Wan2.2-14B~\cite{wan}
& 75.4 & 89.4 & 90.6 & 98.4 & 48.1 & \underline{70.6} & 19.4 & 92.3 & 94.3 \\
DreamXWorld~\cite{dreamxworld}
& 74.8 & 90.3 & 92.2 & 97.3 & 45.6 & 66.2 & 18.9 & 92.4 & 95.8 \\
HunyuanVideo1.5~\cite{hunyunavideo}
& 76.9 & 89.0 & 90.6 & \underline{99.2} & \textbf{51.0} & 70.1 & \underline{19.6} & \textbf{97.9} & 97.8 \\
LingBotWorld~\cite{lingbotworld}
& \underline{77.0} & \underline{91.8} & \underline{93.2} & 99.0 & \underline{50.0} & 66.8 & \textbf{19.7} & \underline{97.8} & \underline{98.0} \\
\textbf{Ours}
& \textbf{77.1} & \textbf{95.4} & \textbf{95.7} & \textbf{99.5} & 44.8 & 67.2 & 19.0 & 96.1 & \textbf{98.8} \\
\bottomrule
\end{tabular}%
}
\end{table}

\subsubsection{General Video Generation}
We evaluate general video generation across complementary settings, covering standard benchmark scenarios as well as synchronized seven-view generation in long-tail and causally structured environments.
Specifically, we report results on the Autonomous Vehicle and Robot subsets of PAI-Bench-G~\cite{paibench}, and complement them with targeted seven-view experiments that probe coherent scene evolution and physical interaction under challenging driving conditions.
PAI-Bench is a comprehensive Physical AI benchmark covering video generation, conditional video generation, and video understanding.
Its generation track, PAI-Bench-G, provides a fine-grained evaluation of perceptual quality, temporal consistency, semantic alignment, reference fidelity, and physical plausibility.

Our evaluation focuses on its Autonomous Vehicle and Robot subsets, which contain 118 and 174 videos, respectively.
These two subsets cover driving-related and embodied-interaction scenarios that align with the primary application domains of our model.
\tabref{tab:av_generation} and \tabref{tab:robot_generation} report the corresponding results.
Our model achieves the highest Quality Score in both domains across all evaluated models, with scores of 78.0 for autonomous driving and 77.1 for robotics.
Compared with Wan2.2-5B, Ours improves Subject Consistency from 90.4 to 96.2 in autonomous driving and from 89.0 to 95.4 in robotics, while also achieving the best Background Consistency, Motion Smoothness, and Image-to-Video Background Consistency in both domains.
These results indicate improved preservation of scene appearance, object identity, temporal continuity, and background structure.
The synchronized seven-view experiments complement this benchmark by examining long-tail content, coherent physical interactions, and cross-view consistency in challenging driving scenarios.

\begin{figure}[!htbp]
    \centering
    \includegraphics[width=\textwidth]{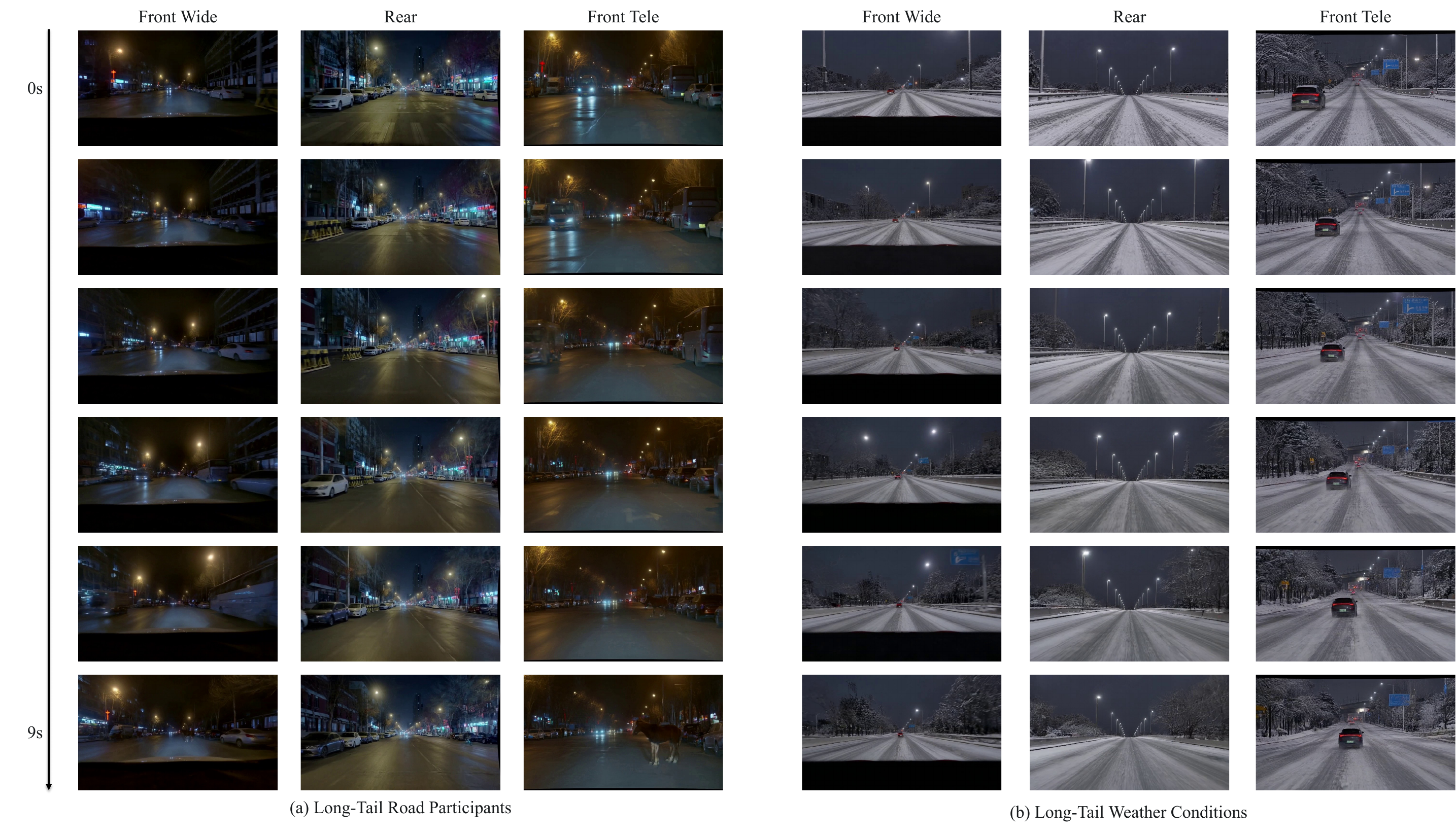}
    \caption{
    Qualitative results for long-tail corner cases.
    For compact visualization, we show the Front Wide, Rear, and Front Tele views selected from the full seven-view output.
    Rows present representative frames from 0 to 9 seconds.
    Left: \emph{Long-Tail Road Participants}, where a horse appears in an urban driving scene.
    Right: \emph{Long-Tail Weather Conditions}, showing nighttime driving on a snow-covered road.
    }
    \label{fig:seven_view_corner_cases}
\end{figure}

\begin{figure}[!htbp]
    \centering
    \includegraphics[width=\textwidth]{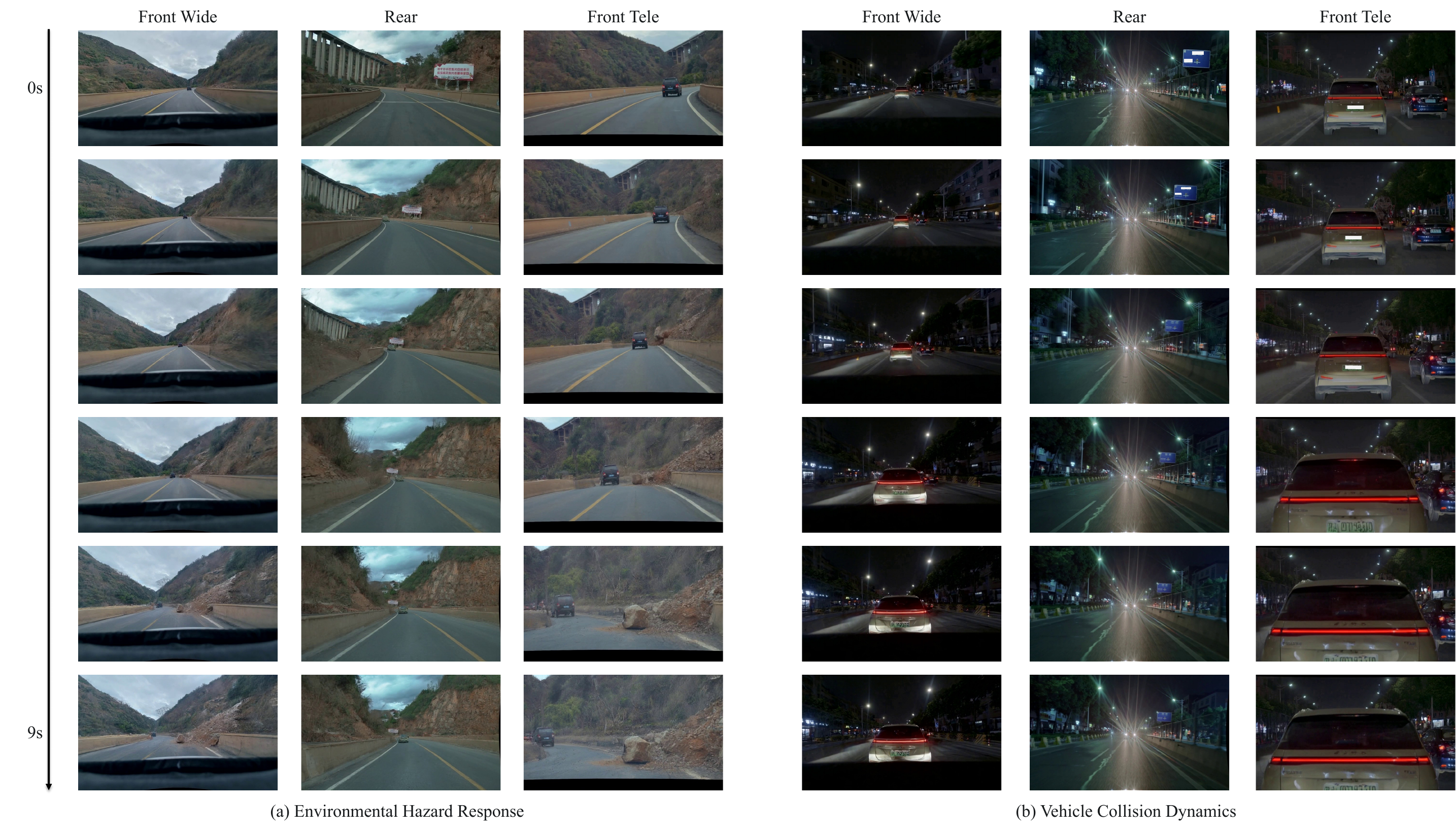}
    \caption{
    Qualitative results for physically causal events.
    Three representative views are shown from the full seven-view output, with rows spanning 0 to 9 seconds.
    Left: \emph{Environmental Hazard Response}, where rocks enter a mountain road and the ego vehicle performs an evasive maneuver.
    Right: \emph{Vehicle Collision Dynamics}, where the ego vehicle progressively approaches and rear-ends the vehicle ahead.
    }
    \label{fig:seven_view_physical_events}
\end{figure}

Beyond PAI-Bench-G, we examine synchronized seven-view generation in two complementary categories: long-tail corner cases that are underrepresented in naturally collected driving data, and physically causal events that require the generated dynamics to remain consistent with real-world physics and preserve coherent cause-and-effect relationships throughout scene evolution.
The generated sequences must preserve a consistent scene state throughout the rollout, with objects and events remaining spatially compatible across views and evolving coherently over time.
For compact visualization, \figref{fig:seven_view_corner_cases} and \figref{fig:seven_view_physical_events} show the Front Wide, Rear, and Front Tele views selected from the full seven-view output.
The rows present representative frames spanning a nine-second generation horizon.
\figref{fig:seven_view_corner_cases} presents two long-tail corner cases.
In the Long-Tail Road Participants example, a horse gradually emerges in the driving scene during the rollout, transitioning from an initially absent object to a newly appearing road participant.
The model incorporates this unexpected participant into the ongoing traffic scene while preserving the temporal stability of the surrounding vehicles, road structure, and background.
The Long-Tail Weather Conditions example depicts nighttime driving on a snow-covered road, where the model maintains the altered road appearance, reduced surface contrast, snow-covered surroundings, and traffic dynamics throughout the sequence.
Together, these examples show that the model can incorporate rare road participants and unusual weather conditions into a temporally coherent scene, while maintaining consistent visual and dynamic changes throughout the rollout.

\figref{fig:seven_view_physical_events} presents two physically causal events designed to examine whether the model can preserve real-world physical consistency and cause-and-effect relationships during scene evolution.
In the Environmental Hazard Response example, rocks fall and roll down the adjacent mountain slope under gravity before entering the roadway, prompting the ego vehicle to perform an evasive maneuver.
The generated sequence preserves the causal order from the initially unobstructed road, to rockfall, and finally to the vehicle's response, while maintaining plausible spatial and temporal relationships among the slope, obstacles, and ego vehicle.
In the Vehicle Collision Dynamics example, the ego vehicle gradually closes the distance to the vehicle ahead, and the rear-end collision occurs only after the two vehicles come into contact.
The continuous reduction in inter-vehicle distance and the subsequent collision reflect a physically plausible interaction rather than an abrupt frame-level change.
Across both examples, object motions and vehicle responses remain consistent with the preceding scene state, while the front-facing views capture the key event progression and the rear view confirms the coherence of the surrounding scene.


\subsubsection{Scene Perception for Video Generation}

A capable world model should generate visually plausible RGB frames while also modeling scene geometry and preserving object-level information over time.
We evaluate these capabilities through depth prediction and referring video segmentation.
During perception-oriented pretraining, RGB videos, depth videos, and referring segmentation masks are represented as video targets and processed by the same latent video generator, without independent task-specific prediction heads.
We assess depth prediction using absolute relative error (AbsRel), where lower is better, and threshold accuracy $\delta_1$, where higher is better.
\tabref{tab:depth_comparison} reports the results on two driving sequences and the reported dataset average.
Our model achieves the strongest average performance among all evaluated methods, with an AbsRel of 0.058 and a $\delta_1$ accuracy of 96.3.
Compared with VGGT-World~\cite{vggtworld}, the strongest competing method, our model reduces AbsRel from 0.065 to 0.058, corresponding to a relative reduction of 10.8\%, and improves $\delta_1$ from 94.0 to 96.3, an absolute gain of 2.3 percentage points.
The advantage is consistent across both driving sequences, where our model obtains the lowest AbsRel and highest $\delta_1$.
It also outperforms the remaining baselines on the reported dataset average, reducing AbsRel by 29.3\%, 39.6\%, and 57.7\% relative to DINO-Foresight~\cite{dinoforesight}, Aether~\cite{zhu2025aether}, and VGGT~\cite{wang2024vggt}, respectively.
These results indicate that the shared video generator learns accurate and transferable geometric representations.

\begin{figure}[ht]
    \centering
    \includegraphics[width=\textwidth]{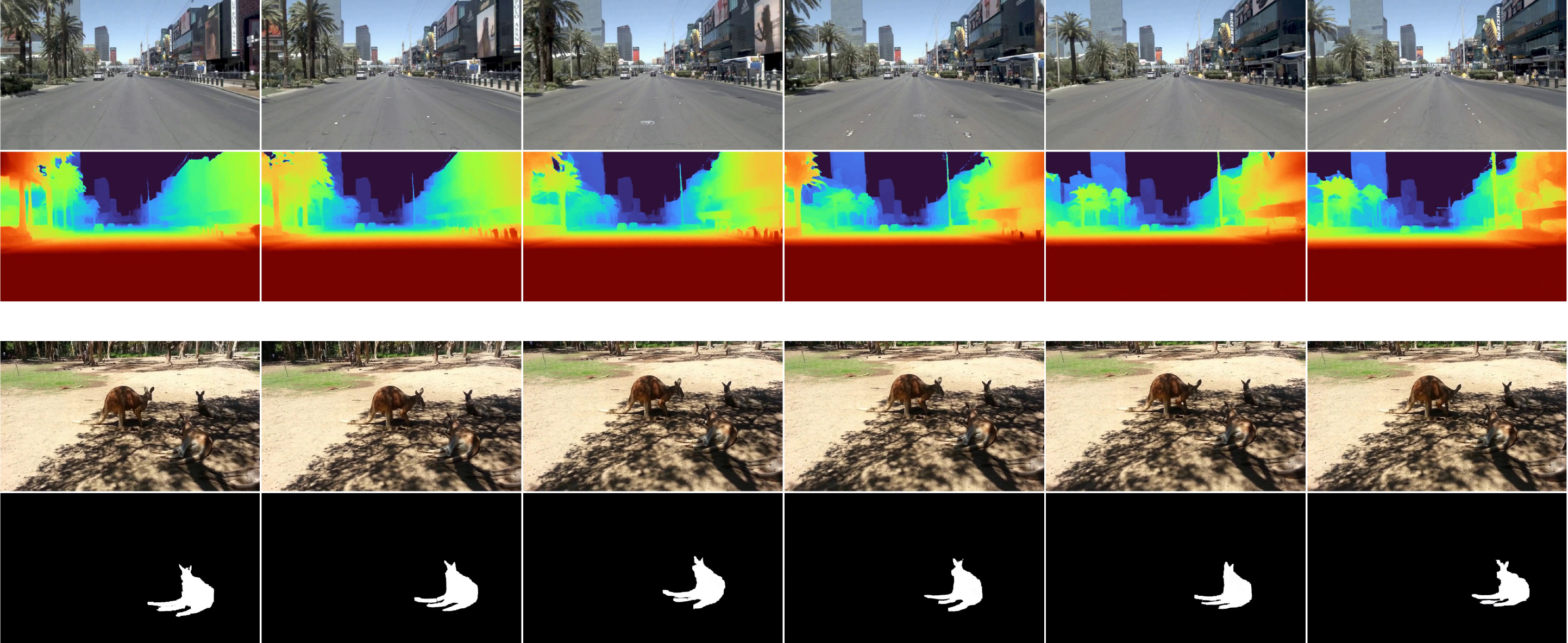}
    \caption{
    Qualitative results for scene perception. Top: RGB frames and corresponding depth predictions. Bottom: RGB frames and referring segmentation masks for target objects. Columns show temporal progression of each sequence.
}
    \label{fig:scene_perception_visualization}
\end{figure}

We also evaluate referring video segmentation to test whether the model preserves object-level information over time.
The model achieves a RefSeg IoU of 0.6726 at reference time $t$.
As the temporal offset increases, IoU remains above 0.55 through $t+10$ (0.5934 at $t+5$ and 0.5501 at $t+10$).
Although performance decreases at larger offsets, the model still obtains IoUs of 0.3104 at $t+20$ and 0.1705 at $t+30$, indicating that object identity, location, visibility, and boundaries remain partially recoverable over extended temporal distances.
Together with the depth comparison, these results show that perception-oriented pretraining provides geometric and object-level temporal priors within the shared video generator.
\figref{fig:scene_perception_visualization} further visualizes both perception tasks.
The predicted depth maps preserve the scene layout and relative depth ordering across frames, while the referring segmentation masks stay aligned with the target object as the scene evolves.

\begin{table}[ht]
\centering
\caption{Depth prediction comparison on two driving sequences and the reported dataset average. AbsRel is lower-is-better, while $\delta_1$ (\%) is higher-is-better. 
}
\label{tab:depth_comparison}
\resizebox{\linewidth}{!}{%
\begin{tabular}{lcccccc}
\toprule
\multirow{2}{*}{Method}
& \multicolumn{2}{c}{\texttt{2011\_09\_26\_drive\_0002\_sync}}
& \multicolumn{2}{c}{\texttt{2011\_10\_03\_drive\_0047\_sync}}
& \multicolumn{2}{c}{Reported Dataset Avg.} \\
\cmidrule(lr){2-3}
\cmidrule(lr){4-5}
\cmidrule(lr){6-7}
& AbsRel $\downarrow$ & $\delta_1 \uparrow$
& AbsRel $\downarrow$ & $\delta_1 \uparrow$
& AbsRel $\downarrow$ & $\delta_1 \uparrow$ \\
\midrule
Cosmos-4B~\cite{Cosmos1} & 0.138 & 84.0 & 0.155 & 77.5 & 0.155 & 77.1 \\
Cosmos-12B~\cite{Cosmos1} & 0.131 & 84.5 & 0.149 & 77.6 & 0.154 & 77.5 \\
Copy Last (VGGT)~\cite{wang2024vggt} & 0.112 & 89.6 & 0.140 & 84.4 & 0.137 & 83.7 \\
Aether~\cite{zhu2025aether} & 0.074 & 94.2 & 0.113 & 87.4 & 0.096 & 87.9 \\
DINO-Foresight~\cite{dinoforesight} & 0.063 & 96.5 & 0.076 & 90.9 & 0.082 & 91.6 \\
VGGT-World~\cite{vggtworld} & \underline{0.050} & \underline{98.3} & \underline{0.064} & \underline{94.9} & \underline{0.065} & \underline{94.0} \\
\textbf{Ours} & \textbf{0.028} & \textbf{99.3} & \textbf{0.059} & \textbf{96.9} & \textbf{0.058} & \textbf{96.3} \\
\bottomrule
\end{tabular}%
}
\end{table}

\subsubsection{Camera-Pose Control for Video Generation}
Camera-pose control evaluates whether explicit geometric conditions can steer the generated viewpoint while preserving scene structure.
Our model uses Fused-PRoPE~\cite{solarwm} to inject camera intrinsics and extrinsics directly into temporal and cross-view attention, allowing the attention computation to account for camera motion and relative view geometry without introducing a separate camera branch.
We adopt the camera controllability metric introduced by WorldScore~\cite{worldscore} for evaluation.
This metric compares the camera trajectory estimated from a generated video with the corresponding input trajectory by combining scale-invariant rotation and translation errors; the resulting error is normalized into a score for which a higher value indicates more accurate camera control.
We compute this metric on two evaluation sets: the static-world subset of WorldScore and a combined set constructed from TartanAir~\cite{wang2020tartanair} and RealCam-Vid~\cite{zheng2025realcam}.
The former measures camera controllability on the benchmark's diverse static scenes, while the latter evaluates generalization to additional scenes and camera trajectories.

\begin{figure}[ht]
    \centering
    \includegraphics[width=\textwidth]{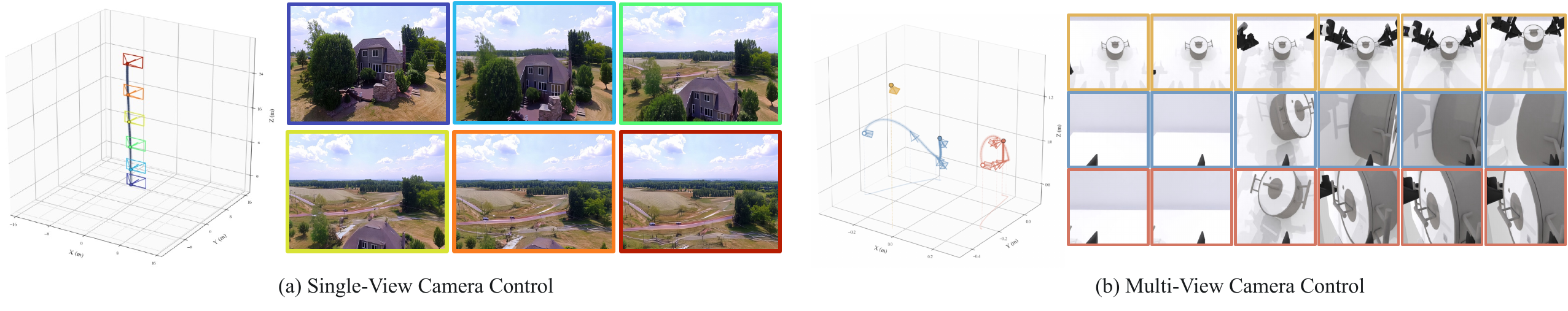}
    \caption{Qualitative results for camera-pose-controlled generation. (a) Single-view camera control, with the prescribed camera trajectory shown on the left and representative generated frames on the right. (b) Multi-view camera control for the head, left, and right cameras, showing their trajectories and the corresponding synchronized observations. The color progression indicates the temporal order from start to end.}
    \label{fig:camera_pose_visualization}
\end{figure}

As reported in \tabref{tab:camera_pose_control}, our model achieves the best camera controllability score on both evaluation sets.
It obtains 86.35 on the WorldScore~\cite{worldscore} static-world subset, exceeding the strongest baseline, FantasyWorld-1.0~\cite{fantasyworld}, by 4.90 points.
On the combined TartanAir~\cite{wang2020tartanair} and RealCam-Vid~\cite{zheng2025realcam} set, our model reaches 70.08 and outperforms FantasyWorld-1.0~\cite{fantasyworld} by 25.80 points.
The consistent improvements across both evaluation sets demonstrate that Fused-PRoPE enables the model to follow specified camera trajectories across diverse scenes and motion patterns.
\figref{fig:camera_pose_visualization} provides qualitative examples in both single-view and multi-view settings.
In the single-view example, the generated frames follow the prescribed camera trajectory while maintaining a stable scene layout.
In the multi-view example, the head, left, and right camera streams follow their respective trajectories and produce mutually consistent observations of the scene.

\begin{table}[t]
\centering
\caption{Camera-pose-controlled video generation evaluated using the camera controllability metric introduced by WorldScore. \textit{WorldScore} denotes the static-world subset of the WorldScore evaluation set,
while \textit{Combined} denotes the evaluation set formed by merging TartanAir and
RealCam-Vid. Higher scores indicate more accurate adherence to the specified camera trajectories. 
}
\label{tab:camera_pose_control}
\begin{tabular}{lccc}
\toprule
\multirow{2}{*}{\centering Method} & \multicolumn{3}{c}{Camera Controllability Score $\uparrow$} \\
\cmidrule(lr){2-4}
& WorldScore & Combined & Average \\
\midrule
CogVideoX-I2V~\cite{cogvideox} & 38.27 & 26.72 & 32.50 \\
Hunyuan 1.5~\cite{hunyunavideo} & 40.15 & 28.71 & 34.43 \\
LingBot World~\cite{lingbotworld} & 42.06 & 35.97 & 39.02 \\
FantasyWorld-1.0~\cite{fantasyworld} & \underline{81.45} & \underline{44.28} & \underline{62.87} \\
DreamXWorld~\cite{dreamxworld} & 81.23 & 43.01 & 62.12 \\
\textbf{Ours} & \textbf{86.35} & \textbf{70.08} & \textbf{78.22} \\
\bottomrule
\end{tabular}
\end{table}

\subsection{Closed-loop Simulation}

In this section, we assess our model which from the Post-train stage as a closed-loop simulator for autonomous driving from three aspects: (i) multi-view driving generation on nuScenes~\cite{Caesar2019nuScenesAM} for benchmarking, (ii) multi-view embodied~\cite{triworldbench2026} results to verify cross-view consistency and action-conditioned scene evolution, and (iii) system-level deployment of our model server to simulate long-horizon, controllable driving scenarios, where the generated observations are consumed by Xiaomi's autonomous driving stack for closed-loop testing.

\subsubsection{Multi-view Driving Results}


We first evaluate on the public nuScenes benchmark for a fair comparison with prior work. Following the standard split, we train on 750 scenes and evaluate on 150 scenes; each scene lasts roughly 20 seconds and contains six synchronized camera views. We report FID and FVD. As shown in \tabref{tab:multiview_driving}, our model achieves competitive performance among short-video generation methods (\eg, Vista~\cite{vista}, DriveDreamer~\cite{drivedreamer}, and MagicDrive-V2~\cite{magicdriveV2}), while supporting generation at arbitrary length. Under the long-video setting, our method outperforms other autoregressive generators (\eg, Self-Forcing(++)~\cite{selfforcing,selfforcing++}, LongLive~\cite{longlive}, and Causal-Forcing++~\cite{causalforcing++}) with the best FVD and the second-best FID, indicating superior temporal quality in driving scenarios; our 6-view results are also competitive.

\begin{figure}[!htbp]
    \centering
    \includegraphics[width=\textwidth]{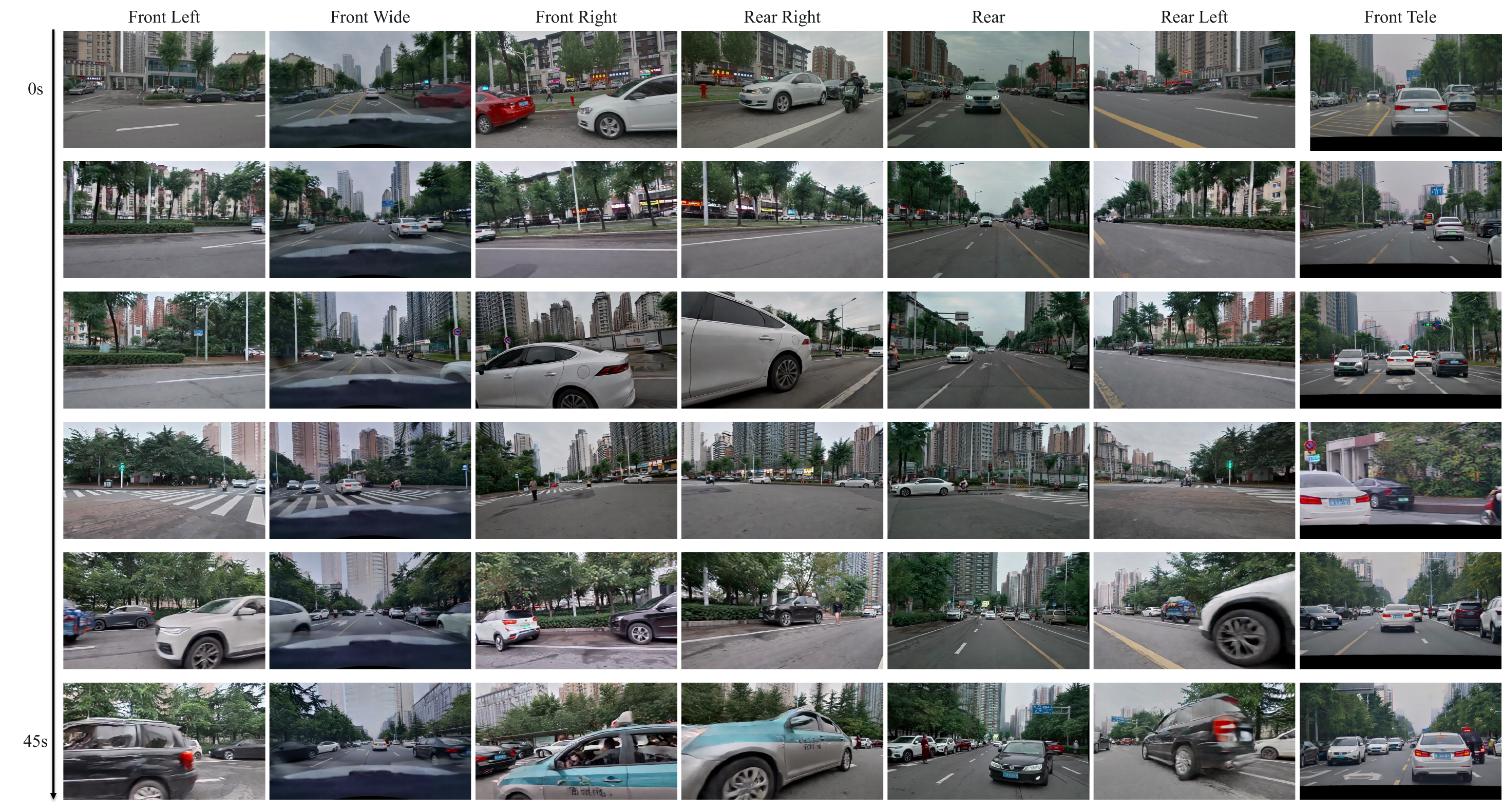}
    \caption{Qualitative results on internal driving data: 7-view long-video generation (45 s). Our model preserves fine-grained structures (\eg, lane markings and traffic signs) over long horizons.}
    \label{fig:mi_internal_45s}
\end{figure}

\begin{figure}[!htbp]
    \centering
    \includegraphics[width=\textwidth]{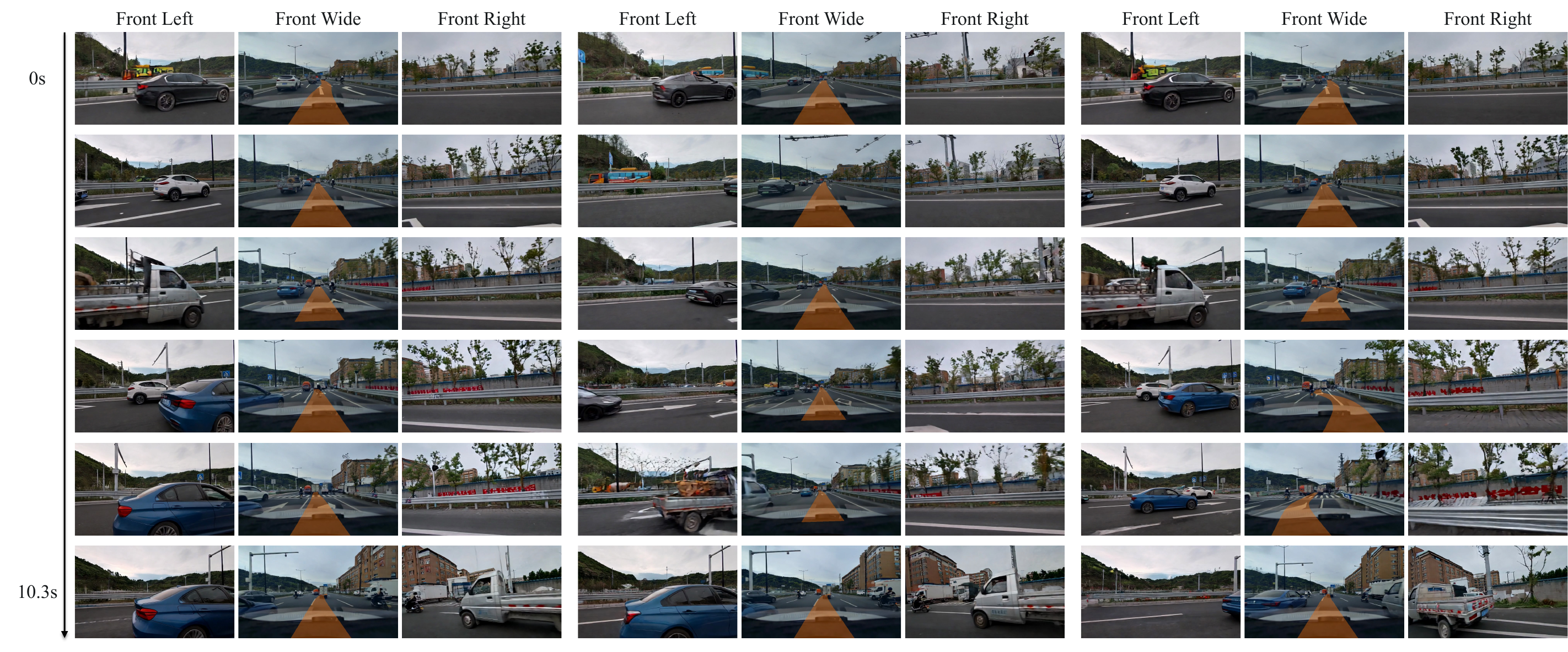}
    \caption{Qualitative results on internal driving data: generation under novel trajectories. Our model remains robust under challenging, out-of-distribution trajectories.}
    \label{fig:mi_internal_3traj}
\end{figure}

We validate on internal driving data. To ensure that the videos remain consistent with the scene, we adopt the rendered reconstructions from \secref{sec:worldrec2} as scene priors for world control. \figref{fig:mi_internal_45s} showcases 7-view long-horizon generation up to 45 seconds, and \figref{fig:mi_internal_3traj} demonstrates generation under diverse, novel trajectories.

\subsubsection{Multi-view Embodied Results}
We evaluate our method in embodied, multi-view settings on TriWorldBench~\cite{triworldbench2026}, an action-conditioned multi-view video generation benchmark. Given a robot action sequence, the first-frame observations from one head camera and two wrist cameras, and a task description, the benchmark requires generating a video sequence that is consistent with the executed actions. The evaluation set contains 500 episodes spanning 50 manipulation tasks. Performance is measured by TWB-Score, which aggregates 19 sub-metrics over six aspects: tri-view consistency, task alignment, 3D coherence, motion quality, temporal consistency, and visual quality. We compare against methods with publicly available papers, and report their numbers from the official leaderboard snapshot dated Sep. 30, 2026.

\begin{table}[ht]
\centering
\caption{Video generation results under short-video and long-video settings.
FID and FVD are lower-is-better. 
}
\label{tab:multiview_driving}
\begin{tabular*}{0.8\linewidth}{@{\extracolsep{\fill}}lcccc@{}}
\toprule
Method & Views & Frames & FID $\downarrow$ & FVD $\downarrow$ \\
\midrule
\rowcolor{gray!15}
\multicolumn{5}{l}{\textit{Short-video Generation}} \\
Vista~\cite{vista} & 1 & 25 & \textbf{6.90} & \textbf{89.40} \\
DriveDreamer~\cite{drivedreamer} & 6 & 8 & 14.90 & 340.80 \\
MagicDrive-V2~\cite{magicdriveV2} & 6 & 16 & 20.91 & \underline{94.84} \\
\multirow{2}{*}{\textbf{Ours}} & 1 & 21 & \underline{12.09} & 135.13 \\
& 6 & 21 & 19.62 & 171.73 \\
\midrule
\rowcolor{gray!15}
\multicolumn{5}{l}{\textit{Long-video generation}} \\
Self-Forcing~\cite{selfforcing} & 1 & 211 & 41.53 & 161.00 \\
Self-Forcing++~\cite{selfforcing++} & 1 & 211 & 28.84 & 147.57 \\
LongLive~\cite{longlive} & 1 & 211 & 29.05 & 161.41 \\
Causal Forcing++~\cite{causalforcing++} & 1 & 211 & \textbf{25.05} & \underline{126.86} \\
\multirow{2}{*}{\textbf{Ours}} & 1 & 211 & \underline{25.57} & \textbf{125.74} \\
& 6 & 211 & 58.90 & 203.76 \\
\bottomrule
\end{tabular*}
\end{table}

\begin{figure}[ht]
  \centering
  \includegraphics[width=\linewidth]{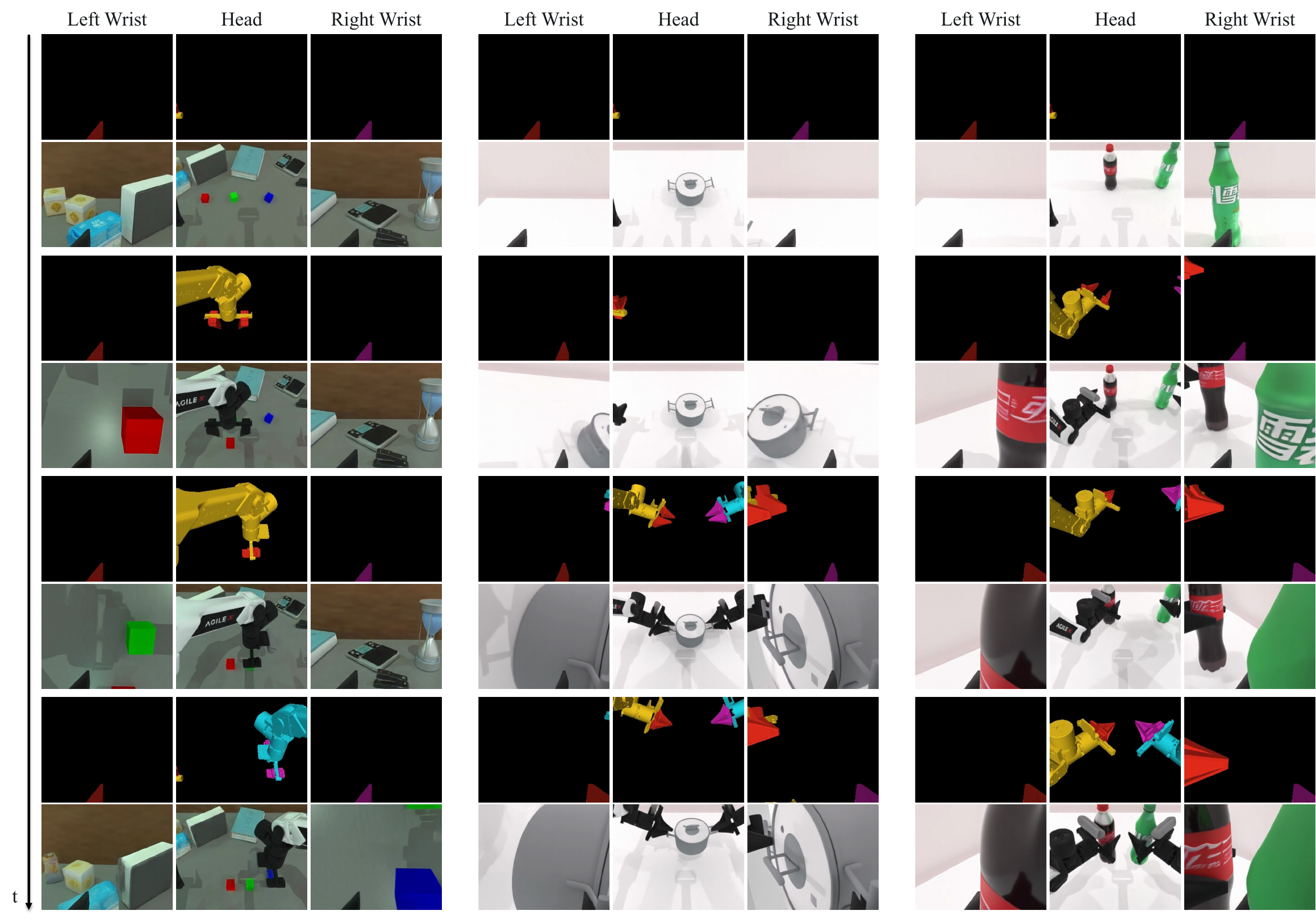}
  \caption{Qualitative visualization of world control (top) and our generated multi-view results (bottom). The world control is rendered from the robot URDF driven by the action sequence.}
  \label{fig:simulation-robot}
\end{figure}

As shown in \tabref{tab:triworldbench}, our approach achieves competitive overall performance without any additional tricks or test-time heuristics, and attains the best Trajectory Accuracy among all compared methods.
\figref{fig:simulation-robot} visualizes representative results. In each row, the top shows the input world control signal, rendered by driving the robot's URDF model with the action sequence. The bottom shows our generated outputs. The results indicate that our model faithfully follows the arm motions specified by the world control, while the camera-view pose control keeps the motion consistent across different cameras.

\begin{table}[!ht]
\centering
\caption{TriWorldBench results on TWB-Score and the metrics where our method ranks first or second among the compared methods.
Bold and underlined scores denote first and second place.
All metrics are higher-is-better.}
\label{tab:triworldbench}
\normalsize 
\setlength{\tabcolsep}{4pt} 
\renewcommand{\arraystretch}{1.15}
\begin{tabularx}{\linewidth}{l *{7}{>{\centering\arraybackslash}X}}
\toprule
Method & TWB-Score & Traj. Acc. & Norm. PSNR & VQA Cons. & Flow Score & Bg. Cons. & Photo. Smooth. \\
\midrule
DreamDojo~\cite{gao2026dreamdojo} & 51.72 & 16.61 & 59.78 & 47.82 & 6.52 & \textbf{89.29} & 15.86 \\
Fysiverse-Video~\cite{wang2026fysiverse} & 63.62 & 37.40 & 68.90 & 63.32 & 22.16 & 75.36 & \textbf{35.63} \\
WoVR\_Plus~\cite{jiang2026wovr} & 65.39 & \underline{45.27} & \textbf{74.62} & 69.11 & \textbf{24.85} & 75.19 & 32.90 \\
BWM~\cite{bwm2026bwm} & \underline{65.54} & 41.97 & 73.43 & 65.94 & 20.48 & 70.29 & 33.95 \\
CausalWM~\cite{causalwm} & \textbf{66.04} & 42.98 & 72.41 & \textbf{70.03} & 22.99 & 75.53 & 30.27 \\
\textbf{Ours} & 64.79 & \textbf{47.25} & \underline{73.44} & \underline{69.31} & \underline{23.76} & \underline{75.57} & \underline{34.16} \\
\bottomrule
\end{tabularx}
\end{table}

\subsubsection{Closed-loop Evaluation Results}
Using the post-trained model together with the deployment optimizations in~\secref{sec:train-infer-infer}, we build a closed-loop evaluation environment to benchmark six versions of our policy model.
Taking hazardous lane change and late deceleration as examples, the relative ranking of these six models in our closed-loop evaluation matches the relative ranking observed in real-vehicle tests.
Specifically, the Pearson correlation is $r=-0.88$ for hazardous lane change and $r=-0.84$ for late deceleration.
These results suggest that the closed-loop evaluation environment built on our Causal World Model is highly reproducible and can be used for policy-model performance evaluation.



\subsection{Closed-loop Training}

In this section, we construct a simulator based on the causal reasoning world model and endow the end-to-end autonomous driving framework with causal reasoning-enabled planning capability through post-training of the end-to-end model. Subsequently, ~\secref{closed_loop_strategy} elaborates the world model-based closed-loop training strategy and experimental setup, and \secref{closed_loop_training_results} reports the corresponding closed-loop training test results.

\subsubsection{Closed-Loop Training Strategy and Setup}
\label{closed_loop_strategy}
\begin{figure}[!htbp]
  \centering
  \includegraphics[width=\textwidth]{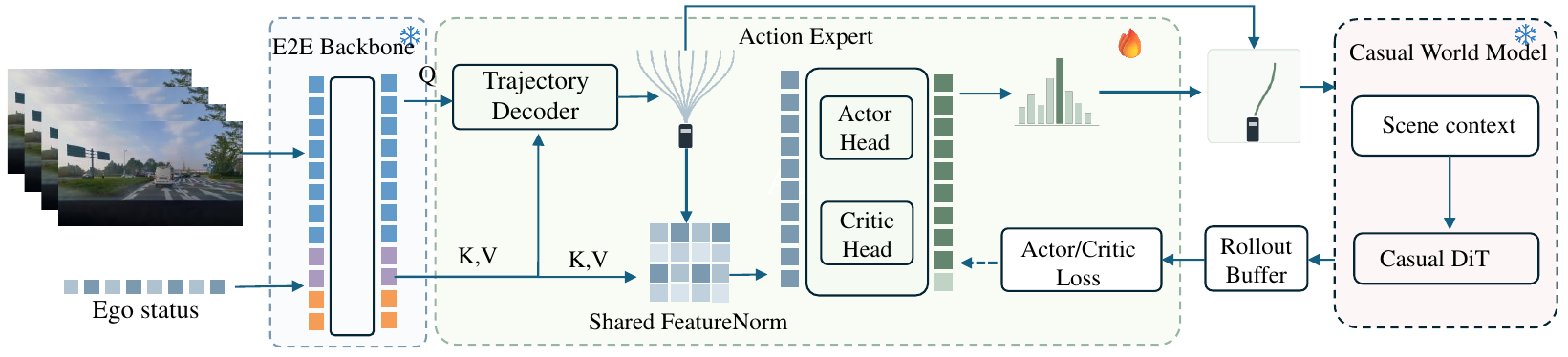}
  \caption{Architecture of the closed-loop training strategy based on the causal reasoning world model simulator.
The strategy comprises three core components: an end-to-end driving model, a reinforcement learning (RL) post-training module, and a causal reasoning world model simulator. Solid lines denote data flow; dashed lines denote gradient propagation. The proposed strategy fine-tunes the action generation model via post-training.}
  \label{fig:clossed_loop_training_framwork}
\end{figure}

At the core of this strategy lies a causal reasoning world model-based simulator, which conducts online post-training of the end-to-end model via closed-loop interaction. This equips the driving policy with causal planning capabilities and improves key performance metrics, including safety and driving efficiency. Specifically, we develop an online RL-enabled post-training scheme based on a state-of-the-art end-to-end driving policy~\cite{Kirby_2026_CVPR,anonymous2026stjepa} to fine-tune its action generation network. \figref{fig:clossed_loop_training_framwork} shows the overall architecture of this closed-loop training strategy, which consists of three modules: the end-to-end driving model, the RL post-training module, and the causal world model simulator.

\vspace{0.2cm}
\noindent \textbf{End-to-End Model and Post-Training Strategy.} Based on a state-of-the-art end-to-end driving model, we redesign the action output by adding a policy head and a value head, making the driving policy compatible with reinforcement-learning post-training. The backbone acts as an observation encoder, mapping raw images, historical trajectories, and ego-state signals into a unified token sequence. The policy predicts trajectories and, through closed-loop interaction with the causal-reasoning world-model simulator, receives future state transitions and reward feedback.
Finally, based on interaction data collected from rollouts, we update the policy network and value evaluation network corresponding to expert actions via the Proximal Policy Optimization (PPO) algorithm. The mathematical formulation of the post-training policy loss function is presented below.
\begin{equation}
    \mathcal{L}(\theta,\phi)=\mathbb{E}\left[\min\big(\rho_t A_t,\text{clip}(\rho_t,1-\epsilon,1+\epsilon)A_t\big)-c_1\big(V_\phi(s_t)-R_t\big)^2+c_2 H(\pi_{\theta}(\cdot|s_t))\right],
    \label{eq:ppo_objective}
\end{equation}
where $\mathcal{L}(\theta,\phi)$ denotes the composite objective function of the PPO algorithm combining the clipped policy surrogate objective, value-function regression loss, and entropy bonus; $\mathbb{E}[\cdot]$ is the expectation taken over sampled environment trajectories; $\rho_t = \pi_\theta(a_t|s_t)/\pi_{\theta_{\text{old}}}(a_t|s_t)$ represents the probability ratio between the updated policy $\pi_\theta$ and the old policy $\pi_{\theta_{\text{old}}}$; $A_t$ is the advantage function at time step $t$; $\text{clip}(\rho_t,1-\epsilon,1+\epsilon)$ is the clipping operator restricting $\rho_t$ within $[1-\epsilon,1+\epsilon]$; $\epsilon$ is the clipping threshold hyperparameter to constrain the magnitude of each policy update; $V_\phi(s_t)$ is the state value predicted by the value network parameterized by $\phi$; $R_t$ denotes the target baseline value for state $s_t$; $c_1$ and $c_2$ are weighting coefficients for the value loss and entropy bonus term, respectively; $H(\pi_{\theta}(\cdot|s_t))$ is the entropy of stochastic policy distribution $\pi_a$, which promotes exploration by discouraging over-deterministic policies; $s_t$ refers to the environment state at time step $t$; and $\pi_{\theta}(\cdot|s_t)$ is the stochastic policy network parameterized by $\theta$.

The advantage function \( A_t \) involved in Eq.~(\ref{eq:ppo_objective}) is computed via generalized advantage estimation (GAE), which is built upon the exponentially weighted accumulation of temporal difference errors. Mathematically, the advantage function is formulated as:
\begin{equation}
A_t = \sum_{l=0}^{T-t-1} (\gamma\lambda)^l \delta_{t+l},   \quad \quad \delta_t = r_t + \gamma V_\phi(s_{t+1}) - V_\phi(s_t),
\end{equation}
\noindent where \( r_t \) denotes the immediate reward received by the agent at time step \( t \); \( \gamma \in [0, 1] \) is the discount factor that weights future rewards; \( V_\phi(s_t) \) and \( V_\phi(s_{t+1}) \) are the predicted state values of states \( s_t \) and \( s_{t+1} \) output by the value network parameterized by \( \phi \), respectively; \( \lambda \in [0, 1] \) serves as the GAE smoothing parameter that regulates the bias-variance trade-off of the advantage estimation; and \( T \) represents the total number of time steps in a single sampled trajectory.

\begin{table}[htbp]
  \centering
  \caption{Test Results of Closed-Loop Training Strategy and End-to-End Baseline Methods. CR denotes collision rate, SCR denotes success rate, OR denotes off-road rate, TTC denotes minimum time-to-collision, EP denotes ego progress rate, and HDR denotes trajectory compliance region rate. The upward arrow $\uparrow$ indicates that higher values correspond to better performance, while the downward arrow $\downarrow$ indicates that lower values correspond to better performance.}
  \setlength{\tabcolsep}{3.5pt} 
  \begin{tabular}{c c c c c c c}
    \toprule
    \multirow{2}{*}{Method} & \multicolumn{4}{c}{Safety} & \multicolumn{2}{c}{Efficiency and Others} \\
    \cmidrule(lr){2-5} \cmidrule(lr){6-7}
    & CR (\%) $\downarrow$ & SCR (\%) $\uparrow$ & OR (\%) $\downarrow$ & TTC (s) $\uparrow$ & EP (\%)$\uparrow$ & HDR (\%) $\uparrow$ \\
    \midrule
    DrivoR-E2E~\cite{Kirby_2026_CVPR} & 1.4 & 80.3 & 0.0 & 0.504 & 81.0 & \underline{99.96} \\
    ST-JEPA-E2E~\cite{anonymous2026stjepa} & \underline{1.2} & \underline{95.2} & 0.0 & \underline{0.614} & \textbf{84.7} & 99.93 \\
    E2E-WM-RL (Ours) & \textbf{0.4} & \textbf{96.3} & 0.0 & \textbf{0.740} & \underline{84.5} & \textbf{99.99} \\
    \bottomrule
  \end{tabular}
  \label{tab:closed_loop_training_result} 
\end{table}

\noindent\textbf{Simulator Based on the Causal Reasoning World Model.} This closed-loop simulator receives future trajectory commands output from the end-to-end model, and estimates the vehicle’s future pose states via its integrated trajectory tracking module and dynamics model. The causal reasoning world model within the simulator takes HD maps, ego vehicle pose, static and dynamic obstacle information, and initial frame data as inputs, generating scene frame information at subsequent timesteps and corresponding state reward signals. Furthermore, the simulator incorporates a Linear Quadratic Regulator (LQR) trajectory tracking controller and a linear two-degree-of-freedom (2-DOF) kinematic model, enabling the mapping from planned trajectories to the vehicle’s future poses. Furthermore, the simulator calculates the total reward \(r_t\) of the current state using the ego-vehicle pose, lane markings, and both dynamic and static obstacles.

\begin{figure}[htbp]
    \centering
    \begin{subfigure}[t]{0.511\textwidth}
        \centering
        \includegraphics[width=\linewidth]{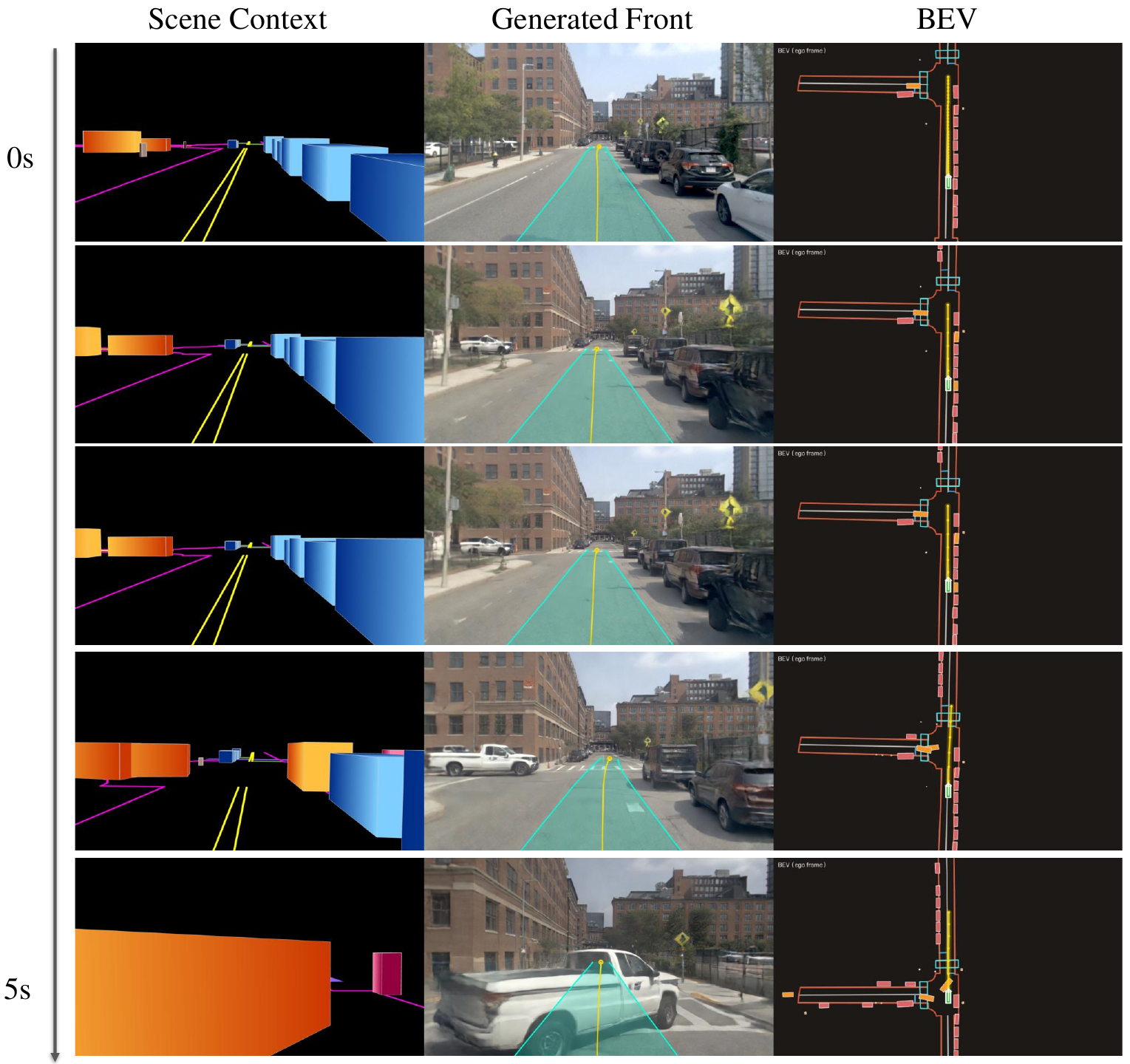}
        \caption{ST-JEPA-E2E}
        \label{fig:closed_loop_view_709_scene_dark_baseline}
    \end{subfigure}
    \hspace{-0.5em}
    \begin{subfigure}[t]{0.482\textwidth}
        \centering
        \includegraphics[width=\linewidth]{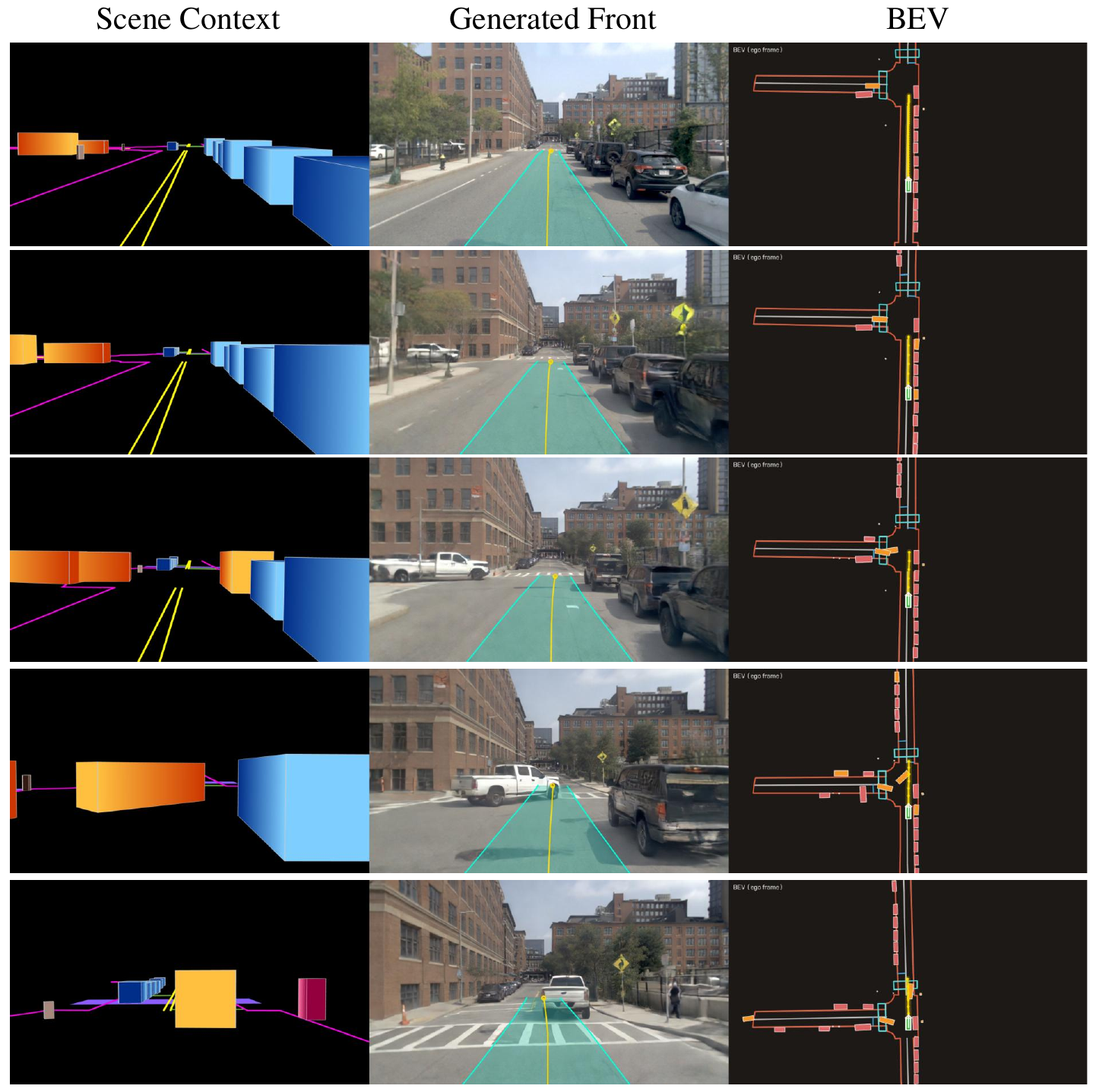}
        \caption{E2E-WM-RL}
        \label{fig:closed_loop_view_709_scene_dark_ours}
    \end{subfigure}
    \caption{Visualization Results of Testing Closed-Loop Training Strategy and End-to-End Policy for Unsignalized Intersection Scenarios. ST-JEPA-E2E is the state-of-the-art end-to-end baseline, and E2E-WM-RL is our RL-posttrained end-to-end policy. Subplots from left to right show the world-model scene control diagram, world-model generated scene with trajectory projections, and BEV visualization with trajectory projections.}
    \label{fig:closed_loop_view_709_scene_dark}
\end{figure}

The total reward is a weighted sum of individual reward terms $r_t=\sum_{i}k_i r_i$
where \(k_i\) denotes the weight coefficient of each reward component, and \(r_i\) represents individual reward terms, consisting of progress reward \(r_{p}\), drivable-area reward \(r_{d}\), Time to Collision (TTC) reward \(r_{tc}\), collision penalty \(r_{c}\), off-road penalty \(r_{o}\), and comfort reward \(r_{a}\). The progress reward \(r_{p}\) characterizes the effective forward travel distance of the vehicle along the driving direction, formulated as \(r_{{p}}=\mathrm{clip}(\Delta p,-5,5)/5\), where \(\Delta p\) is the effective forward distance projected from vehicle displacement onto the heading direction. The drivable-area reward \(r_{d}\) is calculated from the fraction of the four corners of the ego bounding box that lie inside the drivable-area polygon. The off-road penalty \(r_{o}\) is a binary indicator term, which takes the value of \(-1\) when severe off-road driving is detected and \(0\) otherwise. The TTC reward is computed from the minimum time-to-collision \({t}_\text{min}\), and can be compactly written as \(r_{{tc}} = -\max\left(0,\ \min\left(1,\ \frac{3.0-t_{\text{min}}}{1.5}\right)\right)\). The collision penalty \(r_{c}\) is also a binary indicator term, assigned as \(-1\) if geometric overlap between the ego vehicle and polygons of dynamic or static obstacles indicates a collision, and \(0\) otherwise. The comfort reward \(r_{a}\) quantifies riding comfort penalties integrated from the violation degrees of vehicle speed, acceleration, yaw rate and jerk relative to predefined thresholds.


\subsubsection{Closed-Loop Training Experimental Results}
\label{closed_loop_training_results}

In this experiment, we adopt the causal inference world model to construct closed-loop simulation training scenarios using the training and validation scene data provided by NavSim, and build closed-loop simulation test scenarios based on the test-set scenes from NavSim. During the closed-loop testing phase, 1000 representative scenes are selected as evaluation cases. The results are shown in 
\tabref{tab:closed_loop_training_result} and  \figref{fig:closed_loop_view_709_scene_dark}.

\begin{figure}[htbp]
    \centering
    \begin{subfigure}[t]{0.511\textwidth}
        \centering
        \includegraphics[width=\linewidth]{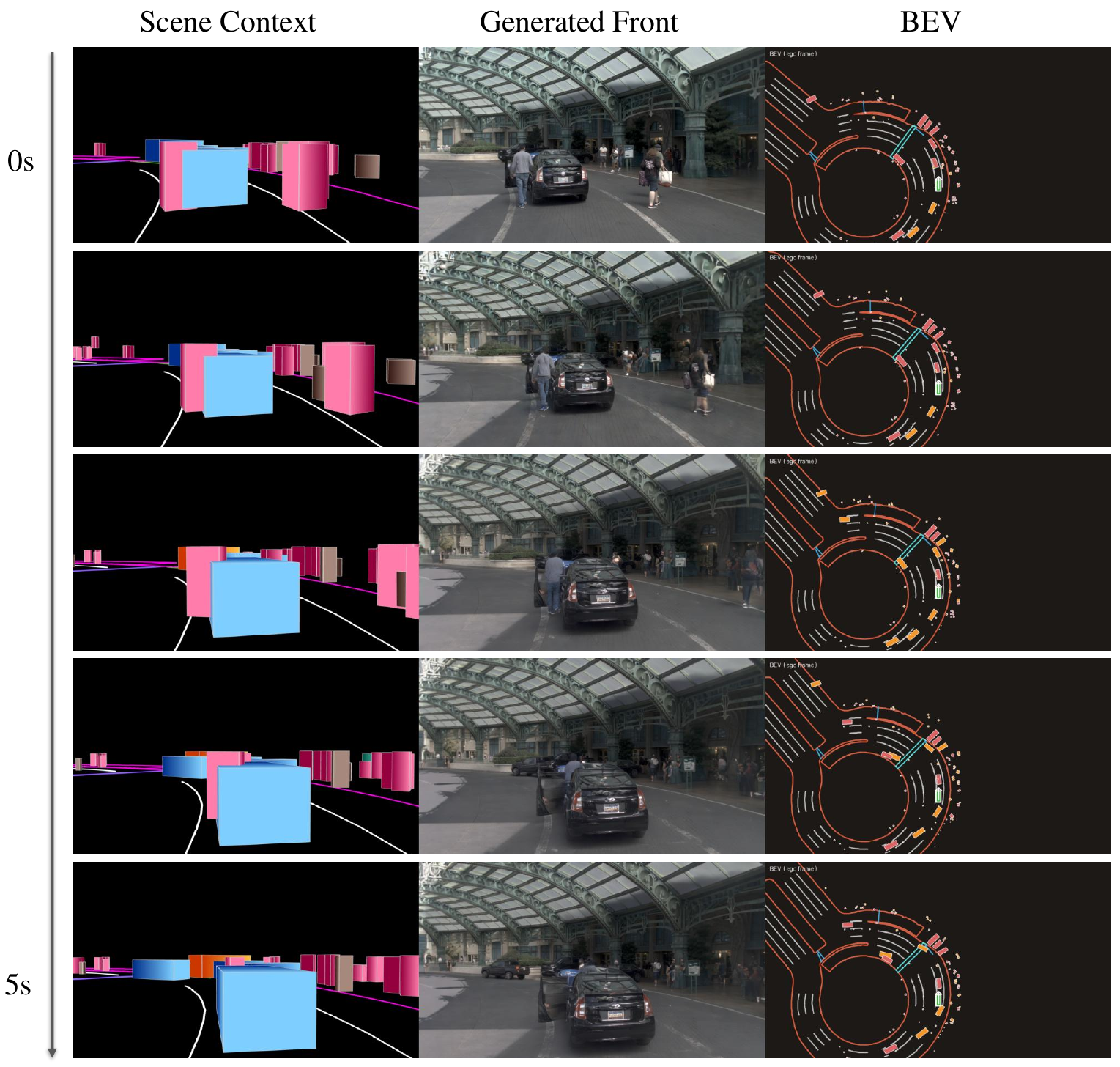}
        \caption{ST-JEPA-E2E}
        \label{fig:closed_loop_view_13_scene_dark_baseline}
    \end{subfigure}
    \hspace{-0.5em}
    \begin{subfigure}[t]{0.482\textwidth}
        \centering
        \includegraphics[width=\linewidth]{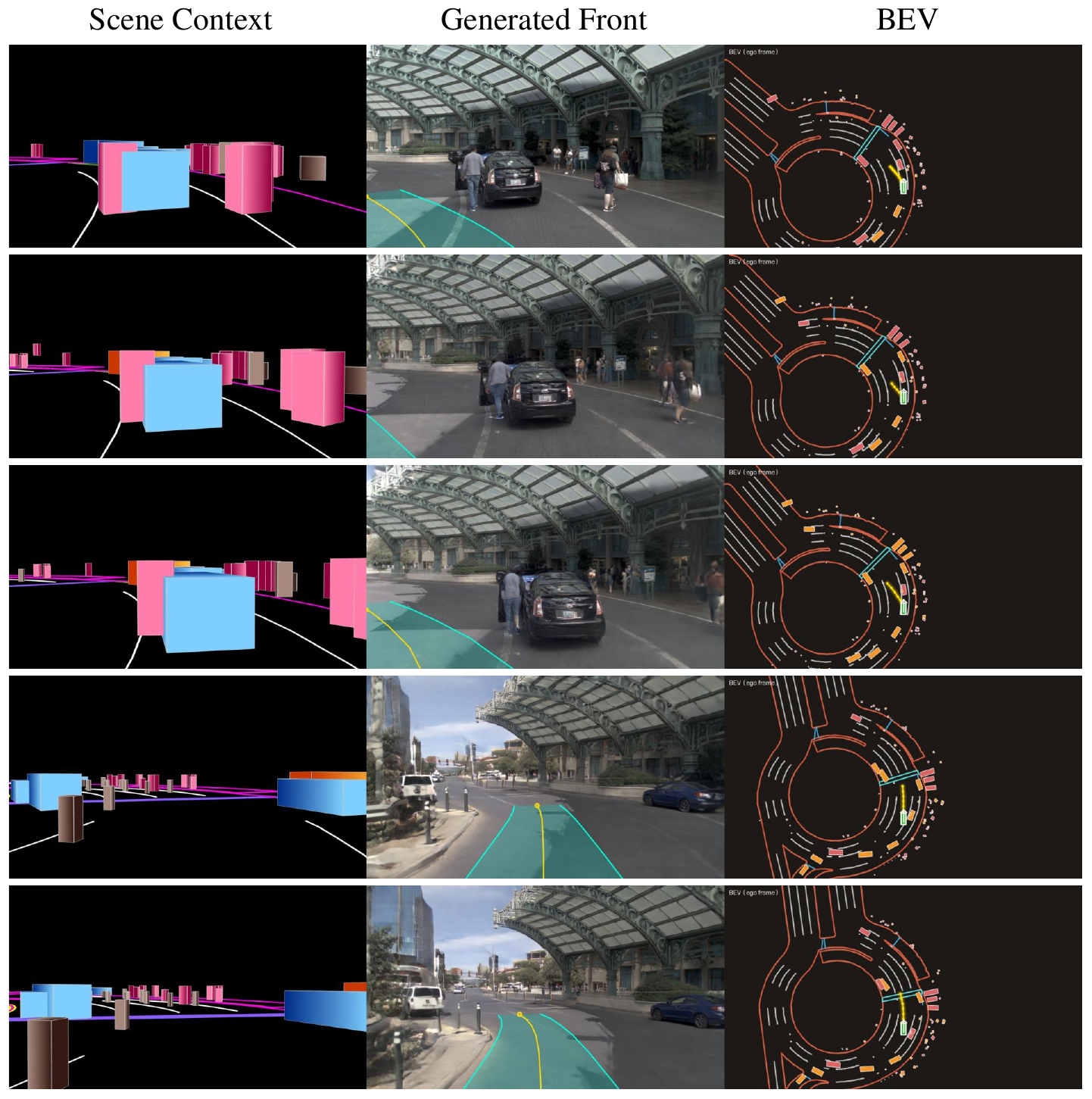}
        \caption{E2E-WM-RL}
        \label{fig:closed_loop_view_13_scene_dark_ours}
    \end{subfigure}
    \caption{Visualization Results of Closed-Loop Training Strategy and End-to-End Policy for Parked Vehicle Occupation Scenarios. ST-JEPA-E2E is the state-of-the-art end-to-end baseline, and E2E-WM-RL is our RL-posttrained end-to-end policy. Subplots from left to right show the world-model scene control diagram, world-model generated scene with trajectory projections, and BEV visualization with trajectory projections.}
    \label{fig:closed_loop_view_13_scene_dark}
\end{figure}


The results of a multi-dimensional performance comparison between state-of-the-art end-to-end baselines and an end-to-end policy built on a causal-reasoning world model architecture with online reinforcement learning fine-tuning are summarized in \tabref{tab:closed_loop_training_result}.
The evaluation metrics span two primary dimensions: driving safety and driving efficiency. The metrics adopted in this work include Collision Rate (CR), Scenario Completion Rate (SCR), Off-Road Probability (OR), TTC, Ego Progress (EP), and Host Driving Region Accuracy (HDR). Among them, CR, SCR, OR, and TTC characterize driving safety performance, while EP and HDR reflect driving efficiency and regulatory compliance, respectively. As shown in \tabref{tab:closed_loop_training_result}, the end-to-end policy E2E-WM-RL derived from closed-loop post-training delivers substantial improvements in collision rate, scenario completion rate, and TTC, effectively elevating the safety performance of the end-to-end framework. In addition, the proposed policy outperforms the baseline in driving region compliance, while retaining competitive performance on the driving efficiency metric EP. These results confirm that interactive closed-loop training enables end-to-end driving policies to achieve improved safety and regulatory compliance while preserving competitive driving efficiency.

A visual comparison of driving behaviors between two end-to-end driving strategies evaluated in a closed-loop simulation environment is presented in \figref{fig:closed_loop_view_709_scene_dark} and \figref{fig:closed_loop_view_13_scene_dark}. The two strategies are the proposed strategy built on a causal inference world model architecture after online closed-loop fine-tuning, and the state-of-the-art end-to-end baseline algorithm.
As depicted in \figref{fig:closed_loop_view_709_scene_dark}, in the intersection merging scenario, the closed-loop-trained E2E-WM-RL strategy can proactively anticipate the motion dynamics of merging vehicles. It completes speed adjustment of the ego vehicle 4 seconds before the merging vehicle enters the conflict zone, which effectively prevents collisions at the merging point. In contrast, the ST-JEPA-E2E end-to-end baseline strategy exhibits a pronounced delay in its deceleration response: it only initiates braking when the merging vehicle approaches the ego vehicle’s driving corridor, and ultimately collides with the merging vehicle during the merging process due to insufficient deceleration. As shown in \figref{fig:closed_loop_view_13_scene_dark}, for the scenario where a temporarily stopped vehicle occupies the forward lane, the E2E-WM-RL strategy actively re-plans its driving trajectory to maneuver around the obstructing vehicle and smoothly traverses the section when a safe passing space is available in the left lane. By comparison, the ST-JEPA-E2E baseline strategy remains in a car-following waiting state and fails to execute active obstacle avoidance maneuvers, eventually resulting in a collision with a trailing vehicle from behind. Overall, the E2E-WM-RL strategy with online closed-loop training can proactively regulate vehicle speed in response to situational changes across driving scenarios, which fundamentally mitigates collision risks and improves driving safety. Meanwhile, it is equipped with dynamic trajectory planning capability, which enhances road throughput efficiency by bypassing lane-occupying obstacles ahead. In summary, the proposed strategy outperforms existing end-to-end baseline algorithms in terms of both driving safety and traffic efficiency.

\subsection{World Action Model Application}
\label{sec:wam_application}

\begin{figure}[!htbp]
  \centering
  \includegraphics[width=\textwidth]{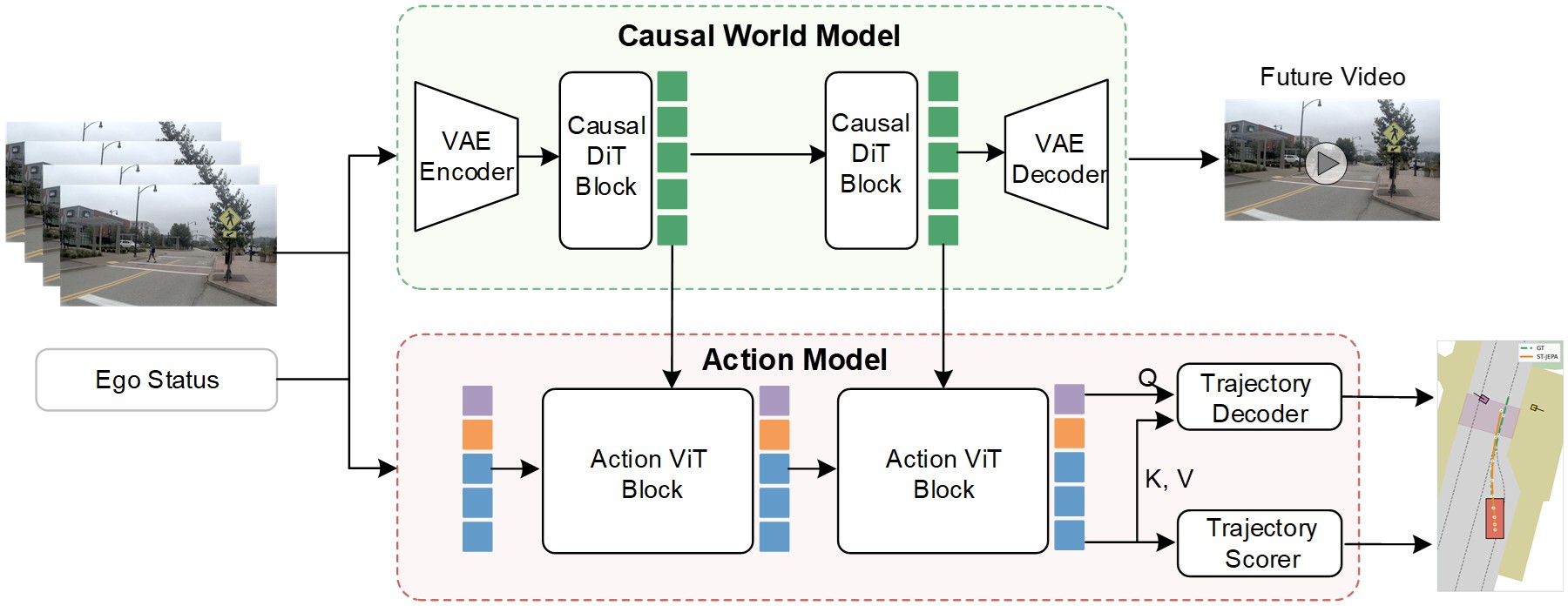}
  \caption{World-Action Model. The causal world model encodes multi-frame observations into a spatiotemporal world latent. The Action Model then combines this latent with multi-camera history and ego status to produce action tokens. These tokens are decoded into candidate future trajectories, which are scored by the PDM heads.}
  \label{fig:WAM_MAIN}
\end{figure}

\subsubsection{World Action Model for Driving }
Generation and reconstruction quality alone do not establish that a world model has learned scene information relevant to driving decisions. Closed-loop driving also requires the model to understand drivable regions, traffic-participant motion, ego status, collision risk, and the effect of ego actions on future states. We therefore connect spatiotemporal world latents from a causal world model to a World-Action Model (WAM) planner and evaluate the interface with on NAVSIM ~\cite{dauner2024navsim} using the Predictive Driver Model Score (PDMS) and its component metrics. 

\begin{figure}[!htbp]
  \centering
  \includegraphics[width=\textwidth]{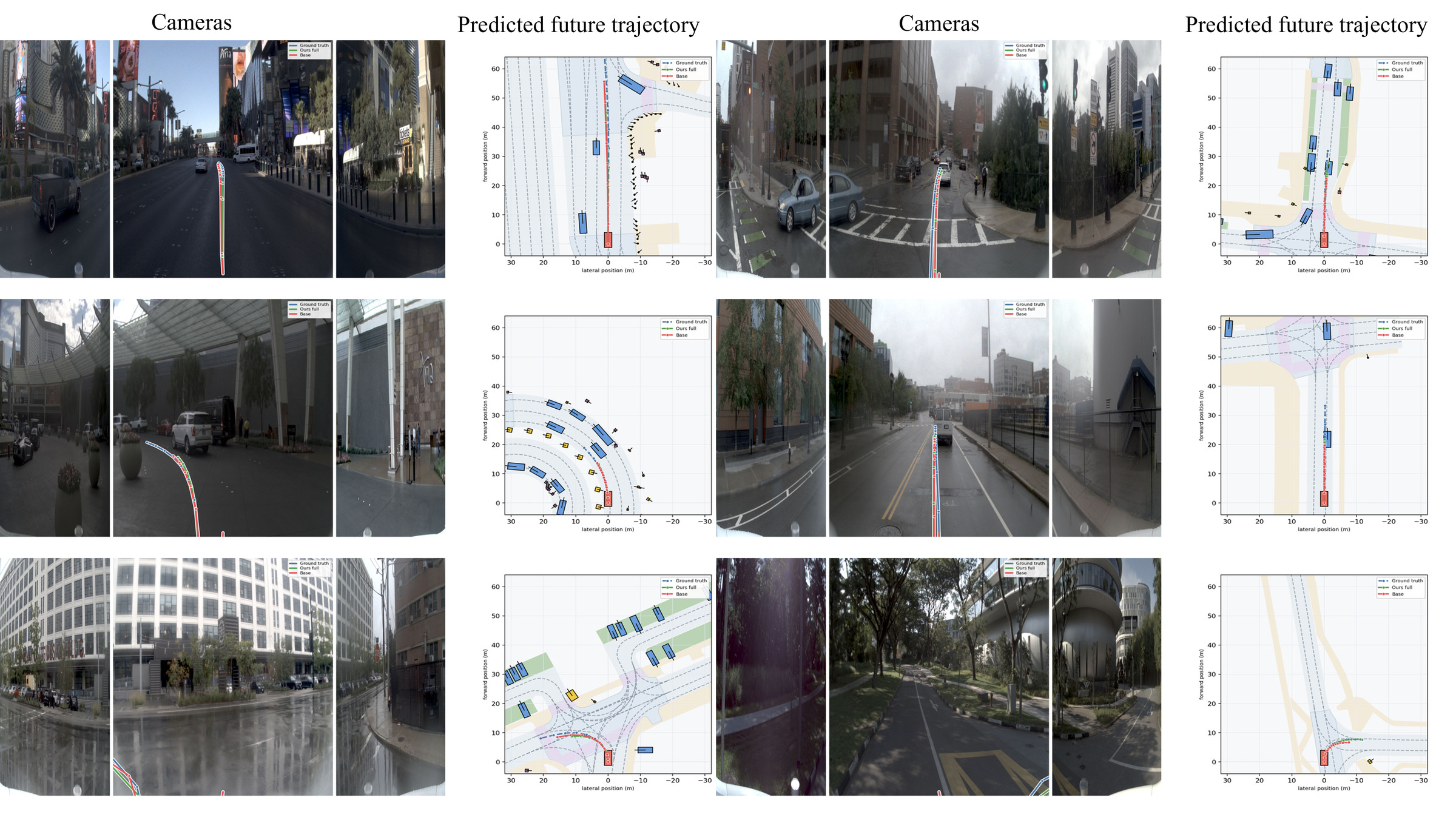}
  \caption{\textbf{Qualitative trajectory comparison for WAM planning.} Candidate trajectories produced by the Action Model conditioned on the causal world model are visualized in the ego-centric scene, showing how the causal world model affects trajectory selection and driving behavior.}
  \label{fig:WAM_TRAJ}
\end{figure}

\paragraph{System structure.}
The system is organized as an upstream causal world model followed by an Action Model planner. The information flow is illustrated in Fig.~\ref{fig:WAM_MAIN} and the qualitative trajectory examples are shown in Fig.~\ref{fig:WAM_TRAJ}. The causal world model receives historical multi-frame observations and encodes a spatiotemporal world latent. It supports an RGB mode and a mode augmented with scene-understanding conditions such as depth and segmentation. This world latent is provided as conditioning information to the Action Model, which also receives multi-camera history from four directions, ego status, and scene conditions. The Action Model maintains action tokens and lets them interact with the world latent to produce updated action representations. A Trajectory Decoder maps the resulting representations to a fixed number of candidate future trajectories, and a scoring module predicts safety, drivability, progress, and comfort terms before selecting the highest-scoring trajectory.

The causal world model input five historical frames from a forward-facing camera. Its backbone contains a spatiotemporal VAE and 33 intermediate DiT layers, which provide representations related to future states and action consequences. The Action Model uses four cameras with four historical frames per camera and a joint DINOv2 ViT-S/14 and VJEPA 2 visual backbone. Each view contains 16 learned scene tokens, and the action branch outputs 64 action tokens. The planning horizon is 4 s with a 0.5 s interval, giving eight time points per candidate trajectory.

\paragraph{Planning validation.}
For the controlled comparison between Base and Base + causal world model, we keep the Action Model, Trajectory Decoder, training budget, data split, and post-processing fixed. Only the use of the world latent and the addition of scene-understanding capabilities are changed. 
Here, NC denotes no-at-fault collision avoidance, DAC denotes drivable-area compliance, TTC denotes time-to-collision, C denotes comfort, and EP denotes ego progress. PDMS is the aggregate Predictive Driver Model Score computed from these behavior terms. In Tab.~\ref{tab:wam_v3_results}, the comparison between Base and Base + causal world model highlights the contribution of the generative world representation. Adding the causal world-model branch increases DAC from 98.7 to 99.0, EP from 92.8 to 93.6, and PDMS from 94.9 to 95.3, while keeping NC and C unchanged. Although TTC changes slightly, the overall PDMS gain of 0.4 percentage points shows that the generated spatiotemporal representation provides useful information for planning. Among the methods listed with NAVSIM Trainval training data, Base + causal world model achieves the highest PDMS of 95.3, exceeding the next-best reported Trainval results of 94.8. This result establishes the best leaderboard performance in the Trainval comparison setting on NAVSIM v1.1 navtest.

\begin{table*}[t]
    \centering
    \caption{WAM planning results on NAVSIM v1.1 navtest. External values are source-reported; dashes indicate unreported or pending submetrics. All metrics are higher-is-better.}
    \label{tab:wam_v3_results}
    \small
    \resizebox{\textwidth}{!}{%
    \begin{tabular}{llrrrrrr}
        \toprule
        Method & Training data & NC $\uparrow$ & DAC $\uparrow$ & TTC $\uparrow$ & C $\uparrow$ & EP $\uparrow$ & \textbf{PDMS} $\uparrow$ \\
        \midrule
        Epona~\cite{zhang2025eponaautoregressivediffusionworld} & Trainval 103k & 97.9 & 95.1 & 93.8 & \underline{99.9} & 80.4 & 86.2 \\
        DriveLaW~\cite{drivelaw} & Trainval 103k & 99.0 & 97.1 & 96.7 & \textbf{100.0} & 81.3 & 89.1 \\
        DriveVLA-W0~\cite{li2026drivevla} & Trainval 103k & 98.7 & \underline{99.1} & 95.3 & 99.3 & 83.3 & 90.2 \\
        DriveFine ~\cite{dang2026drivefine} & Trainval 103k & 98.6 & 97.9 & 95.2 & \underline{99.9} & 85.5 & 90.7 \\
        R2SE~\cite{dang2026drivefine} & Trainval 103k & 99.0 & 97.9 & 96.4 & \textbf{100.0} & 86.8 & 91.6 \\
        iPad~\cite{guo2026ipad} & Trainval 103k & 98.6 & 98.3 & 94.9 & \textbf{100.0} & 88.0 & 91.7 \\
        SparseDriveV2~\cite{sun2026sparsedrivev2} & Trainval 103k & 98.5 & 98.4 & 95.0 & \underline{99.9} & 88.6 & 92.0 \\
        Centaur~\cite{sima2026Centaur} & Trainval 103k & \textbf{99.5} & 98.9 & \textbf{98.0} & \textbf{100.0} & 85.9 & 92.6 \\
        DriveSuprim~\cite{yao2026Drivesuprim} & Trainval 103k & 98.6 & 98.6 & 95.5 & \textbf{100.0} & 91.3 & 93.5 \\
        DA-WAM~\cite{zhong2026wam} & Trainval 103k & 99.1 & 98.9 & 96.8 & 99.8 & 90.0 & 93.7 \\
        DrivoR~\cite{Kirby_2026_CVPR} & Trainval 103k & 99.0 & 98.9 & 96.7 & \textbf{100.0} & 90.0 & 93.7 \\
        Drive-JEPA~\cite{Wang2026drivejepa} & Trainval 103k & -- & -- & -- & -- & -- & 93.7 \\
        ChainFlow-VLA~\cite{Wang2026ChainFlow-VLA} & Trainval 103k & \underline{99.2} & 99.0 & \underline{97.2} & \underline{99.9} & 91.9 & 94.8 \\
        FPD-Drive~\cite{chen2026fpd} & Trainval 103k & -- & -- & -- & -- & -- & 94.8 \\
        DriveZero~\cite{he2026drivezero} & Trainval 103k & 99.0 & \textbf{99.2} & 96.0 & \textbf{100.0} & \underline{93.1} & 94.8 \\
        \midrule
        \textbf{Base} & Trainval 103k & 99.1 & 98.7 & 97.0 & \underline{99.9} & 92.8 & \underline{94.9} \\
        \textbf{Base + causal world model} & Trainval 103k & 99.1 & 99.0 & 96.6 & \underline{99.9} & \textbf{93.6} & \textbf{95.3} \\
        \bottomrule
    \end{tabular}}
\end{table*}

\begin{table}[H]
\caption{\textbf{LIBERO Plus} Benchmark results (\% success rate). Avg: weighted average.}
\label{tab:libero_overall}
\begin{center}
\resizebox{\linewidth}{!}{
\begin{tabular}{l c c c c c c c c c}
\toprule
Method
& Embodied PT.
& Layout
& Camera
& Robot
& Language
& Light
& Background
& Sensor
& Avg \\
\midrule
Fast-WAM~\cite{Fastwam} & No
& 61.4 & 16.5 & 44.5 & 67.0 & 79.9 & 52.6 & 38.2 & 49.9 \\
Fast-WAM-Joint~\cite{Fastwam} & No & 74.8 & 27.9 & 60.9 & 89.3 & 86.7 & 49.3 & 51.1 & 62.2 \\
4D-WAM~\cite{4dwam} & Yes & \underline{79.2} & \underline{45.2} & 64.3 & \underline{90.6} & 94.3 & 57.7 & 69.1 & 71.0 \\
GAM & Yes & 79.1 & \textbf{83.1} & \underline{70.0} & 84.8 & \underline{97.2} & \textbf{94.3} & \textbf{95.3} & \textbf{85.5} \\
\textbf{Ours} & No 
& \textbf{81.5} & 39.6 & \textbf{74.7} & \textbf{95.3} & \textbf{97.5} & \underline{71.0} & \underline{71.4} & \underline{75.0} \\
\bottomrule
\end{tabular}
}
\end{center}
\end{table}

\subsubsection{World Action Model for Robot }

Concretely, we instantiate a World Action Model (WAM) for embodied control on top of our Causal World Model, following Fast-WAM~\cite{Fastwam}. We reuse the Causal World Model DiT as the video-branch backbone, and introduce an additional action-expert DiT that shares attention with the video DiT. During training, the video DiT takes the clean first-frame latents together with future noisy latents, while action tokens in the action DiT are restricted to attend only to the clean first-frame latents. At inference time, the video DiT performs a single forward pass conditioned on the clean first-frame latents. We initialize the video DiT with the bidirectional model weights after mid-training.

\tabref{tab:libero_overall} presents our LIBERO Plus~\cite{Liberoplus} benchmark results, where our method reaches a 75.0\% average success rate. It surpasses Fast-WAM~\cite{Fastwam} and Fast-WAM-Joint~\cite{Fastwam}, with the most prominent gains observed under layout, camera, and background distribution shifts. These results indicate that the representations learned by our Causal World Model during pre-training and mid-training transfer effectively to downstream WAM, providing strong spatial priors for embodied manipulation planning.

\subsection{WorldRec 2.0 Reconstruction Experiments}
\label{sec:worldrec2_experiments}

\subsubsection{Evaluation Setup}
This section evaluates the reconstruction method described in \secref{sec:worldrec2}. The experiments measure novel-view appearance and geometry reconstruction and compare WorldRec 2.0 with representative state-of-the-art reconstruction methods under a common protocol. Results are reported on the in-distribution Waymo benchmark in \secref{sec:worldrec2_waymo} and under cross-dataset transfer in \secref{sec:worldrec2_zero}. We also report the results on the MiData in \secref{sec:worldrec2_midata}.

\begin{figure}[!htbp]
  \centering
  \includegraphics[width=\textwidth]{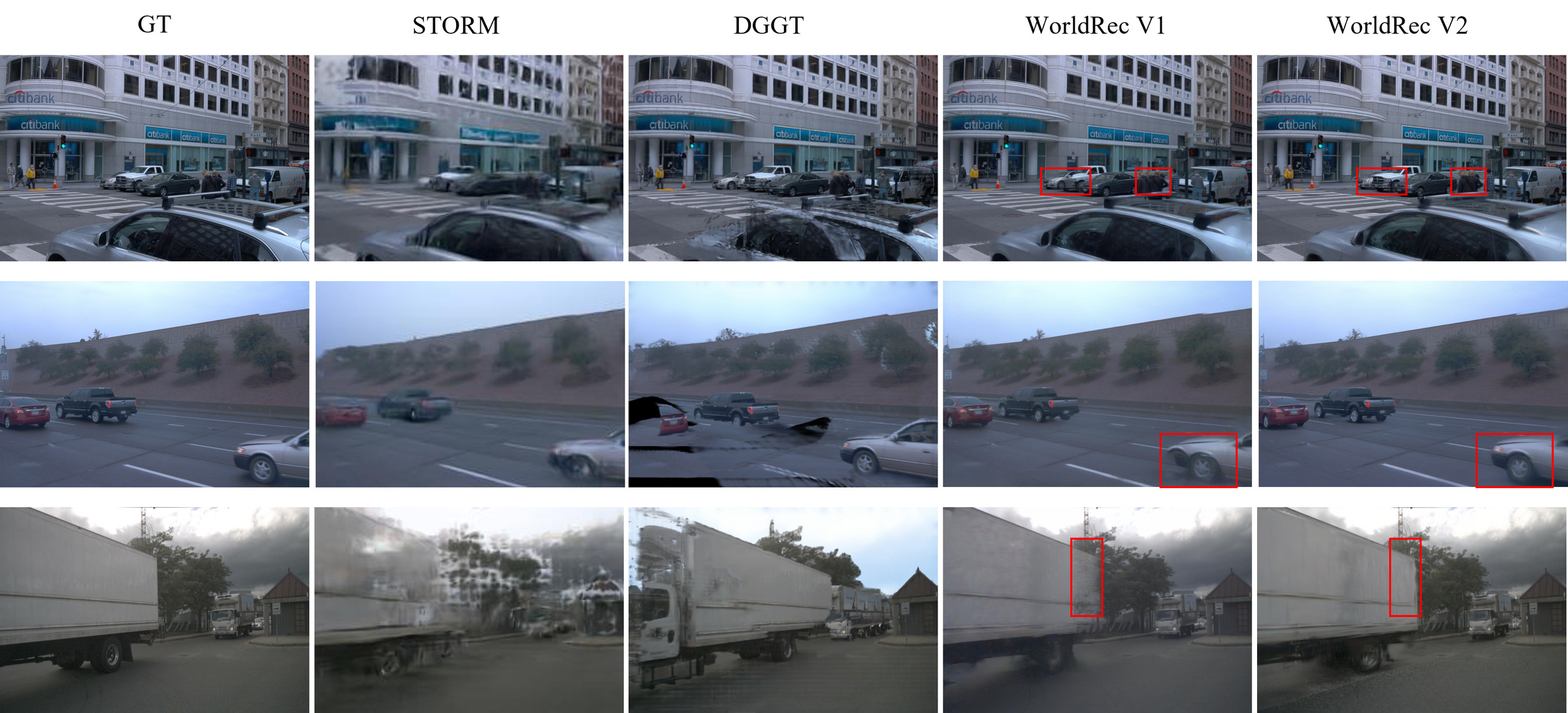}
  \caption{Qualitative results on Waymo and nuScenes. WorldRec 2.0 produces geometry-consistent RGB renderings from unseen camera views and improves the reconstruction of dynamic regions.}
  \label{fig:WorldRec2_waymo}
\end{figure}

Experiments use Waymo Open Dataset ~\cite{xu2026wod} for the main comparison and nuScenes ~\cite{Caesar2019nuScenesAM} and Argoverse 2 ~\cite{Wilson2023Argoverse2} for cross-dataset evaluation. The model uses 20-frame clips from three frontal cameras, four context views, an ImageNet-pretrained EfficientNetV2-S encoder, and InfiniDepth input depth. We report PSNR, SSIM, and depth RMSE (D-RMSE), with higher PSNR/SSIM and lower D-RMSE indicating better reconstruction.

\begin{figure}[!htbp]
  \centering
  \includegraphics[width=\textwidth]{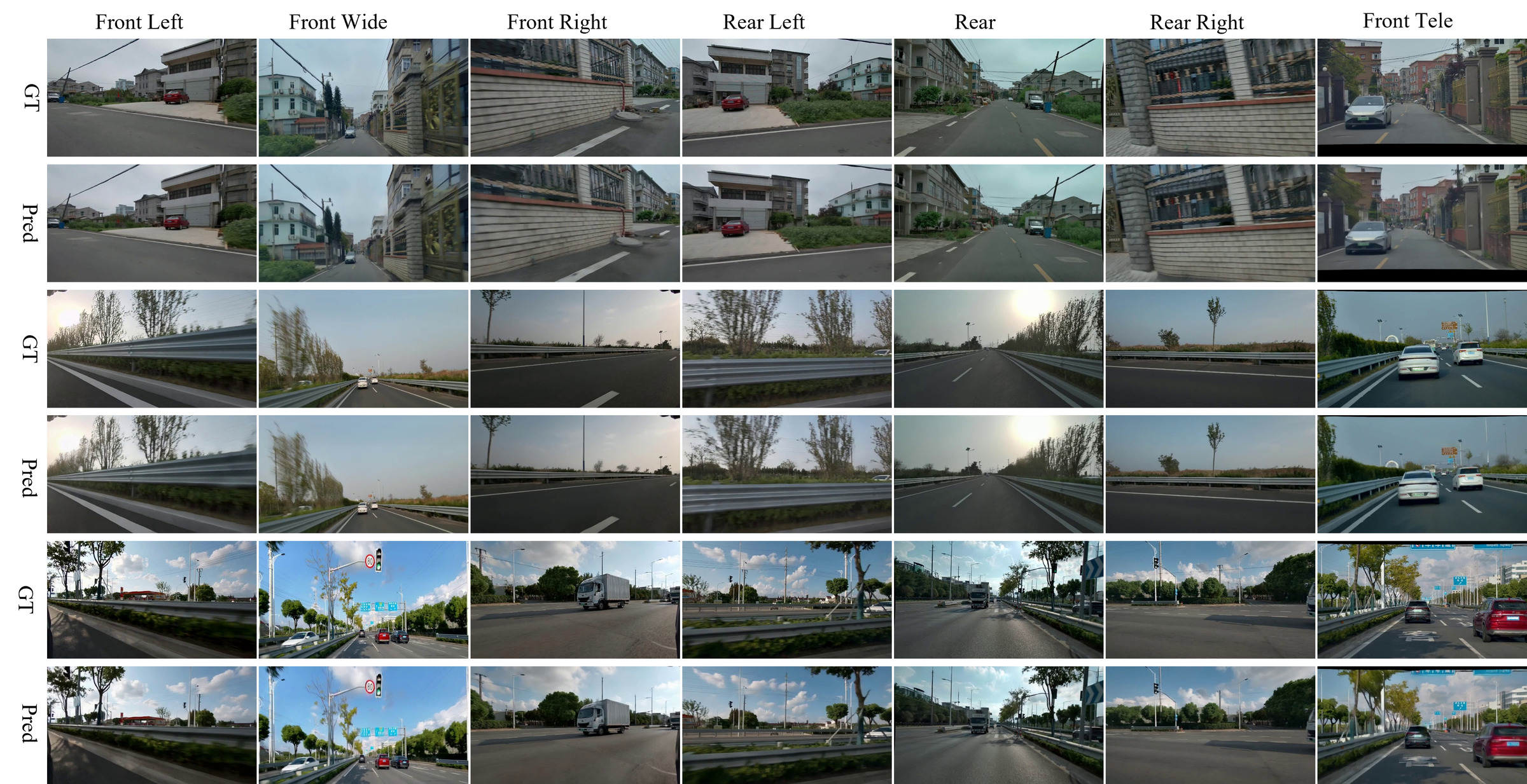}
  \caption{Qualitative 7V reconstruction on the MiData. }
  \label{fig:WorldRec2_MIDATA1}
\end{figure}

\begin{figure}[H]
  \centering
  \includegraphics[width=\textwidth]{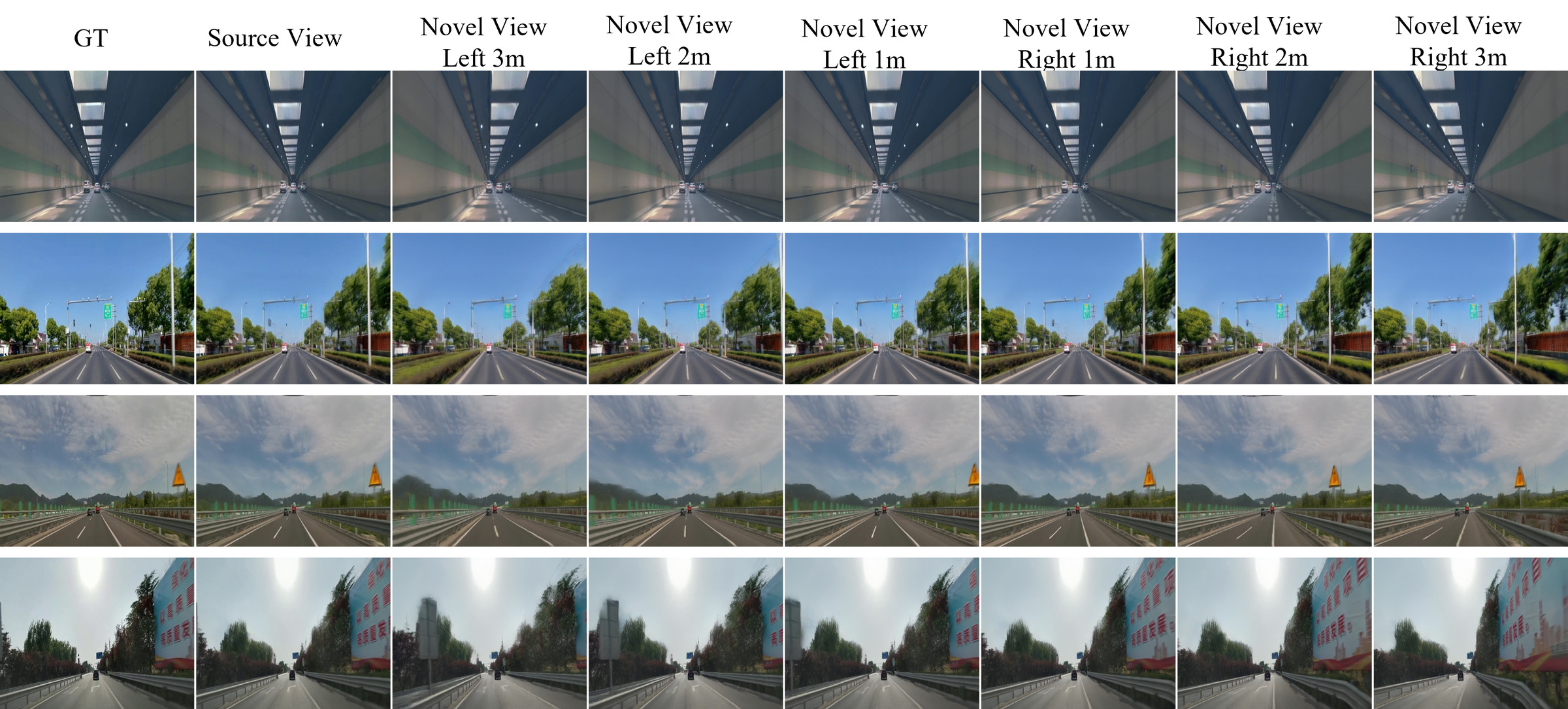}
  \caption{Qualitative novel view results on the MiData.}
  \label{fig:WorldRec2_MIDATA2}
\end{figure}

\begin{table*}[!ht]
    \centering
    \caption{Novel-view reconstruction on Waymo. Dynamic metrics are
    evaluated inside dynamic-object regions; full metrics use the complete
    image.}
    \label{tab:worldrec2_waymo_full}
    \small
    \resizebox{\textwidth}{!}{%
    \begin{tabular}{lrrrrrr}
        \toprule
        Method & Dyn. PSNR $\uparrow$ & Dyn. SSIM $\uparrow$
        & Dyn. D-RMSE $\downarrow$ & Full PSNR $\uparrow$
        & Full SSIM $\uparrow$ & Full D-RMSE $\downarrow$ \\
        \midrule
        EmerNeRF ~\cite{Yang2024Emernerf} & 17.79 & 0.255 & 40.88 & 24.51 & 0.738 & 33.99 \\
        3DGS ~\cite{kerbl20233d} & 17.13 & 0.267 & 13.88 & 25.13 & 0.741 & 19.68 \\
        PVG ~\cite{chen2026periodic} & 15.51 & 0.128 & 15.91 & 22.38 & 0.661 & 13.01 \\
        DeformableGS ~\cite{Yang2024Deformablegs} & 17.10 & 0.266 & 12.14 & 25.29 & 0.761 & 14.79 \\
        LGM ~\cite{Tang2024lgm} & 19.58 & 0.443 & 9.43 & 23.59 & 0.691 & 8.02 \\
        GS-LRM ~\cite{zhang2025Gs-lrm} & 20.02 & 0.520 & 9.95 & 25.18 & 0.753 & 7.94 \\
        Depth Anything 3 ~\cite{lin2026depth} & 15.96 & 0.429 & 12.20 & 23.27 & 0.686 & 11.66 \\
        AnySplat ~\cite{jiang2025anysplat} & 15.99 & 0.418 & 10.45 & 23.32 & 0.687 & 10.31 \\
        STORM ~\cite{yang2025storm} & 22.10 & 0.624 & 7.50 & 26.38 & 0.794 & 5.48 \\
        DGGT ~\cite{chen2025dggt} & 22.80 & 0.651 & 6.55 & 27.41 & 0.846 & 3.47 \\
        WorldRec 1.0 & \underline{25.01} & \underline{0.768} & \underline{6.07} & \underline{28.48} & \underline{0.861} & \underline{3.35} \\
        \textbf{WorldRec 2.0} & \textbf{25.81} & \textbf{0.773}
        & \textbf{3.99} & \textbf{29.88} & \textbf{0.865}
        & \textbf{2.89} \\
        \bottomrule
    \end{tabular}}
\end{table*}

\subsubsection{Quantitative Reconstruction Results on open datasets}
\paragraph{Waymo novel-view reconstruction.}
\label{sec:worldrec2_waymo}
WorldRec 2.0 achieves the strongest full-image and dynamic-region metrics
under the common protocol in Tab.~\ref{tab:worldrec2_waymo_full}. Relative to WorldRec 1.0, full-image PSNR increases
by 1.40 dB and D-RMSE decreases from 3.35 to 2.89. Dynamic D-RMSE decreases
from 6.07 to 3.99, a relative reduction of approximately 34.3\%. The larger
geometric improvement in dynamic regions supports the role of dynamic
instance registers, canonical alignment, and intra-instance temporal
interaction.

\paragraph{Cross-Dataset Generalization}
\label{sec:worldrec2_zero}
In the zero-shot setting Tab.~\ref{tab:worldrec2_cross_dataset_full}, WorldRec 2.0 improves over 1.0 by 1.83 dB on
nuScenes and 0.99 dB on Argoverse 2. After fine-tuning, the gains remain
1.72 dB and 0.97 dB. These results show that the method is not specific to Waymo: it maintains
strong novel-view reconstruction accuracy under changes in camera layout and
data distribution, demonstrating improved cross-dataset generalization.

\begin{table*}[t]
    \centering
    \caption{Cross-dataset novel-view reconstruction.}
    \label{tab:worldrec2_cross_dataset_full}
    \small
    \begin{tabular}{llrrrr}
        \toprule
        \multirow[c]{2}{*}{Setting} & \multirow[c]{2}{*}{Method} & \multicolumn{2}{c}{nuScenes}
        & \multicolumn{2}{c}{Argoverse 2} \\
        \cmidrule(lr){3-4}\cmidrule(lr){5-6}
        & & PSNR $\uparrow$ & SSIM $\uparrow$
        & PSNR $\uparrow$ & SSIM $\uparrow$ \\
        \midrule
        Zero-shot & STORM ~\cite{yang2025storm} & 17.77 & 0.669 & 20.83 & 0.542 \\
        Zero-shot & DGGT ~\cite{chen2025dggt} & 25.31 & 0.794 & 25.02 & 0.801 \\
        Zero-shot & WorldRec 1.0 & \underline{26.54} & \underline{0.821} & \underline{25.50} & \underline{0.810} \\
        Zero-shot & \textbf{WorldRec 2.0} & \textbf{28.37} & \textbf{0.829}
        & \textbf{26.49} & \textbf{0.824} \\
        \midrule
        Fine-tuning & STORM ~\cite{yang2025storm} & 24.54 & 0.784 & 24.97 & 0.791 \\
        Fine-tuning & DGGT ~\cite{chen2025dggt} & 26.63 & 0.813 & 26.96 & 0.831 \\
        Fine-tuning & WorldRec 1.0 & \underline{27.50} & \underline{0.826} & \underline{27.66} & \underline{0.851} \\
        Fine-tuning & \textbf{WorldRec 2.0} & \textbf{29.22} & \textbf{0.847}
        & \textbf{28.63} & \textbf{0.872} \\
        \bottomrule
    \end{tabular}
\end{table*}

\subsubsection{Qualitative Reconstruction Results on MiData}
\label{sec:worldrec2_midata}

To further assess reconstruction quality on in-house driving data, we visualize
WorldRec 2.0 on the MiData dataset. MiData provides synchronized
multi-camera observations from seven views. Fig.~\ref{fig:WorldRec2_MIDATA1}
shows the reconstructed seven-view scene, while Fig.~\ref{fig:WorldRec2_MIDATA2}
shows a novel-view reconstruction. These examples qualitatively demonstrate
that the same explicit scene representation supports both multi-view
reconstruction and rendering from a target view beyond public benchmarks.


\phantomsection
\addcontentsline{toc}{section}{Contributors}
\section*{Contributors}
\label{sec:contributors}

\noindent
\textbf{Data:} Zhenxin Zhu $\cdot$ Zhijian Huang $\cdot$ Mingfei Tu $\cdot$ Yonggang Li $\cdot$ Yuru Wang $\cdot$ Jianhua Wu $\cdot$ JiaKui Hu $\cdot$ Ruiqiang Zhang  \\[4pt]
\textbf{Generation Model:} Zhenxin Zhu $\cdot$ Zhijian Huang $\cdot$ Yuru Wang $\cdot$ Jianhua Wu $\cdot$ Yonggang Li $\cdot$ Jiayi Guan $\cdot$ JiaKui Hu $\cdot$ Tao Xu\\[4pt]
\textbf{WorldRec Model:} Cheng Chi $\cdot$ Xianqi Wang $\cdot$ Mingfei Tu $\cdot$ Yuechuan Pu $\cdot$ Hao Li $\cdot$ Qizeng Jia \\[4pt]
\quad \textbf{Video Generation Evaluation:} Zhijian Huang $\cdot$ Jianhua Wu $\cdot$ Yuru Wang $\cdot$ Yonggang Li $\cdot$ Xuhan Sheng $\cdot$ Jian Mao $\cdot$ Ruiqiang Zhang $\cdot$ Yu Zhu \\[4pt]
\quad \textbf{Closed-loop Simulation:} Zhenxin Zhu $\cdot$ Hao Li $\cdot$ Qizeng Jia $\cdot$ Yuechuan Pu $\cdot$ Tao Xu $\cdot$ Ruiling Yang $\cdot$ Zhijian Huang\\[4pt]
\quad \textbf{Closed-loop Training:} Jiayi Guan $\cdot$ Cheng Chi $\cdot$ Zhenxin Zhu $\cdot$ Jian Mao \\[4pt]
\quad \textbf{World Action Model:} Cheng Chi $\cdot$ Hongxiao Yu $\cdot$ Jiayi Guan $\cdot$ Zhenxin Zhu \\[4pt]
\textbf{Project Advisor:} Guang Chen $\cdot$ Hangjun Ye \\[4pt]
\textbf{Project Lead:} Bing Wang
\section*{Acknowledgements}
\label{sec:acknowledgements}
We are deeply grateful to the Cosmos team~\cite{Cosmos3} for their helpful advice in data processing and model deployment.
We sincerely thank the following colleagues for their valuable contributions and support:
Zehan Zhang $\cdot$
Shikun Huang $\cdot$
Lei Gong $\cdot$
Kaixin Xiong $\cdot$
Lei Zhang $\cdot$
Fangzhen Li $\cdot$
Jingyang Zhao $\cdot$
Xi Zhou $\cdot$
Chitian Sun $\cdot$
Chenming Wu $\cdot$
Jiong Deng $\cdot$
Hongwei Xie $\cdot$
Ming Lu $\cdot$
Kun Ma.

{
    \small
    \bibliographystyle{ieeenat_fullname}
    \bibliography{paper}
}

\end{document}